\documentclass[a4paper,fleqn]{cas-sc}

\usepackage[numbers,sort&compress]{natbib}

\usepackage{amsmath,amssymb}
\usepackage{booktabs}
\usepackage{tabularx}
\usepackage{makecell}
\usepackage{array}
\usepackage{graphicx}
\usepackage{adjustbox}
\usepackage{longtable}
\usepackage{xcolor}
\usepackage{ulem}
\newif\ifreviewversion
\ifdefined\ReviewVersion\reviewversiontrue\else\reviewversionfalse\fi
\newenvironment{revision}{\begingroup\ifreviewversion\color{red}\fi}{\par\endgroup}
\newcommand{\rev}[1]{{\ifreviewversion\color{red}\fi #1}}

\newcolumntype{Y}{>{\raggedright\arraybackslash}X}

\newcommand{\inlinefig}[4]{%
  \begin{center}
    \refstepcounter{figure}\label{#1}%
    \includegraphics[width=#2]{#3}\\[4pt]
    {\sffamily\small Figure~\thefigure: #4}
  \end{center}
}
\newcommand{\inlinetable}[3]{%
  \begin{center}
    \begin{minipage}{\linewidth}
    \centering
    \refstepcounter{table}\label{#1}%
    {\sffamily\small Table~\thetable: #2}\\[4pt]
    #3
    \end{minipage}
  \end{center}
}

\newcommand{\tablenote}[1]{\par\vspace{3pt}{\footnotesize\raggedright\textit{Note.} #1\par}}

\newcounter{algorithm}

\newcommand{\inlinealgorithm}[3]{%
  \begin{center}
    \begin{minipage}{\linewidth}
    \refstepcounter{algorithm}\label{#1}%
    {\sffamily\small Algorithm~\thealgorithm: #2}\\[4pt]
    #3
    \end{minipage}
  \end{center}
}

\begin{document}
\let\WriteBookmarks\relax
\def\floatpagepagefraction{1}
\def\textpagefraction{.001}

\rev{\shorttitle{Confidence-Based Materials Data Quality Assessment}
\shortauthors{Sihan Liang et al.}}

\title[mode=title]{Confidence-Based Materials Data Quality Assessment via Domain Knowledge and Graph-Structured Learning}

\author[1,2]{Sihan Liang}
\author[1,2]{Ji Zhang}
\author[3]{Kehan Wang}
\author[1]{Zihao Gao}
\author[3]{Wei Ma}
\author[1,2]{Yangang Wang}
\author[1,2]{Zongguo Wang}[orcid=0000-0002-7719-761X]
\cormark[1]
\ead{wangzg@cnic.cn}
\credit{Conceptualization, Methodology, Software, Writing -- original draft}

\affiliation[1]{organization={Computer Network Information Center, Chinese Academy of Sciences},
            city={Beijing},
            postcode={100083},
            country={China}}
\affiliation[2]{organization={University of Chinese Academy of Sciences},
            city={Beijing},
            postcode={101408},
            country={China}}
\affiliation[3]{organization={School of Materials and New Energy, Ningxia University},
            city={Yinchuan},
            postcode={750021},
            country={China}}

\cortext[1]{Corresponding author}

\begin{abstract}
The rapid development of materials machine learning relies on large-scale, heterogeneous datasets, yet model reliability is often compromised by imperfect data from diverse experiments, simulations, and database integration. Real-world materials databases may contain measurement errors, failed calculations, missing or duplicate records, and hidden violations of physicochemical constraints. Such problems are difficult to identify because corrupted samples may remain statistically plausible rather than appearing as explicit outliers.
In this paper, we propose a confidence-based materials data quality assessment framework that integrates domain knowledge with graph-structured learning. Data quality is formulated as a continuous confidence estimation problem rather than a binary anomaly identification task, enabling sample-level assessment and historical-reference-based batch comparison. 
A nine-dimensional feature-quality prior and duplicate-aware sample initialization are calibrated using materials domain rules and statistical evidence. Confidence is further refined through sample-feature bipartite graph propagation and weakly supervised graph attention, while a Memory-GAT module provides historical reference scoring. Experiments on 23 ODDS benchmark datasets show that the generic module achieves an average AUC-ROC of 0.866. In five repeated mixed-contamination experiments on mp\_20, rule calibration improves HistGradientBoostingRegressor test $R^2$ from $0.538\pm0.008$ to $0.616\pm0.014$. Graph refinement increases average precision from $0.943\pm0.004$ to $0.953\pm0.004$, while the rule-calibrated configuration has the highest mean test $R^2$ among the evaluated screening configurations. Five repeated MACE subset experiments further show an increase in $R^2$ from $0.590\pm0.053$ to $0.880\pm0.027$ after rule-based screening. The framework provides sample- and feature-level confidence with inspectable evidence for materials data review and downstream modeling.
\end{abstract}


\maketitle

\section{Introduction}
Data-driven machine learning has emerged as a fundamental paradigm in materials science, enabling large-scale exploration of composition-structure-property relationships and accelerating the transition from empirical trial-and-error toward data-driven materials design~\citep{himanen2019data}. However, model performance is determined not only by algorithmic architectures but also by the reliability, representativeness, and scientific consistency of the training data. Low-quality data can distort learned structure-property relationships, reduce model generalization, and introduce unreliable predictions into subsequent materials screening and design workflows. Therefore, systematic data quality assessment and continuous quality control are essential for developing reliable and trustworthy artificial intelligence (AI) systems in materials science.

Despite the success of several world-class materials databases, including Materials Project~\citep{jain2013commentary}, OQMD~\citep{kirklin2015open}, AFLOWLib~\citep{calderon2015aflow}, NOMAD~\citep{draxl2018nomad}, and the Materials Data Facility~\citep{blaiszik2016materials}, maintaining consistent and reliable data quality remains challenging due to the diversity of computational workflows, experimental measurements, and curation procedures involved. 
Nevertheless, records in these databases may contain unconverged density-functional theory calculations~\citep{ain2024precise}, failed structural relaxations, unreasonable crystal structures, inconsistent units or field semantics, missing key properties, duplicate entries, and violations of chemical or physical constraints. 
Importantly, many of these issues cannot be detected by conventional statistical criteria because corrupted samples may remain within physically plausible distributions while violating underlying scientific constraints.

Existing data quality assessment and anomaly detection methods~\citep{samariya2023comprehensive}, including multidimensional data quality frameworks~\citep{carvalho2025data}, distance- and density-based detectors, ensemble outlier methods~\citep{rahmati2024ardod}, reconstruction-error models, and deep representation-learning approaches for tabular anomaly detection~\citep{wang2020machine}, have provided valuable tools for identifying corrupted records in generic datasets. 
However, their direct application to materials science data remains challenging due to three fundamental limitations. 
First, many approaches focus on a single quality dimension or binary anomaly labels, rather than constructing a unified and interpretable quality representation across samples, features, and datasets~\citep{schmidt2019recent}. Second, purely statistical methods are sensitive to distribution shapes and feature scales, and their performance can be degraded in sparse and high-dimensional materials descriptor spaces~\citep{wani2025advancing}. Third, many low-quality materials records arise from violations of physical constraints, chemical relationships, structural consistency, or experimental conditions rather than simple numerical deviations.

To incorporate scientific knowledge into data quality evaluation, recent studies have explored knowledge-assisted approaches that combine domain rules with statistical analysis~\citep{li2023interpreting}. 
Knowledge-assisted methods demonstrate that expert rules can improve scientific consistency evaluation and provide inspectable explanations for low-confidence records~\citep{liu2025domain}. 
However, these approaches still lack a unified framework that integrates domain knowledge, statistical evidence, and relational structures among data records for continuous, interpretable, and comparable quality assessment. 
In addition, materials science data are often submitted in batches, sources, or projects, making relative confidence scores within a single dataset insufficient for cross-batch quality comparison and acceptance~\citep{yu2026cognitive}.

This paper proposes a confidence-based materials data quality assessment framework via domain knowledge and graph-structured learning, where data quality is formulated as a continuous confidence estimation problem rather than a binary anomaly detection task. 
The framework first constructs a domain-informed quality prior and then refines confidence through heterogeneous graph learning, including global sample-feature consistency propagation, local neighborhood refinement, and historical reference-based quality assessment. An overview of the proposed framework is presented in Fig.~\ref{fig:abstract_graph}. 
The quality concepts summarize the assessment goals; the numerical indicators and calibration rules are specified in Sections~3.1--3.2.

\begin{center}
    \refstepcounter{figure}
    \includegraphics[
        width=0.92\linewidth,
        keepaspectratio
    ]{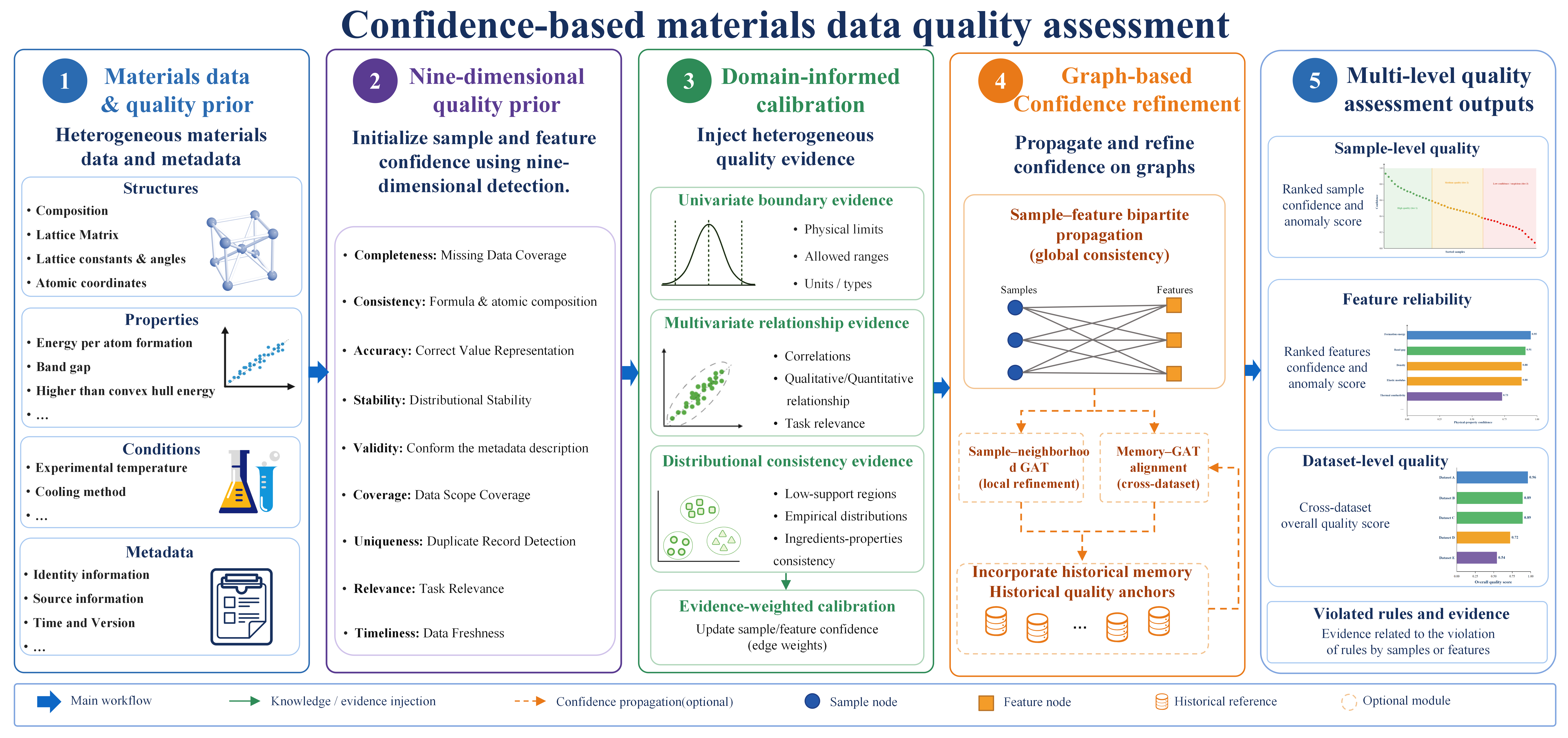}

    \vspace{0.5em}

    \textbf{Figure \thefigure.}
    Overview of the proposed confidence-based materials data quality assessment framework. 
    The five stages connect materials data, quality priors, domain evidence, graph refinement and assessment outputs.
    
    \label{fig:abstract_graph}
\end{center}

The main contributions of this work are summarized as follows:
\begin{enumerate}
\item We formulate materials data quality assessment as a continuous confidence estimation problem, enabling interpretable sample and feature assessment beyond binary anomaly identification;
\item We develop a domain-informed quality prior and hierarchical calibration strategy that integrates physicochemical constraints with statistical evidence;
\item We introduce graph-based confidence refinement at the sample--feature and sample-neighborhood levels, together with historical reference scoring, and evaluate their contributions separately from rule calibration;
\item We evaluate the framework through generic anomaly benchmarks, repeated controlled-contamination and downstream prediction experiments, coefficient sensitivity analysis, and an interpretable materials-data case study. 
\end{enumerate}

\section{Related work}
\subsection{Materials data quality assessment}
Data quality assessment has evolved from single-dimensional statistical checks to multidimensional reliability evaluation frameworks. The “fitness for use” perspective emphasizes that data quality should be evaluated according to accuracy, completeness, consistency, accessibility, and task relevance~\citep{rangineni2023review}. With the increasing adoption of machine learning in scientific discovery, data quality has become closely associated with model reliability, uncertainty, and downstream generalization capability~\citep{priestley2023survey}. 
However, most existing data quality frameworks were originally developed for information systems or industrial data governance, where quality dimensions mainly focus on structural integrity, operational consistency, and data management requirements rather than scientific validity~\citep{kokott2024towards}.

In materials science, large-scale data infrastructures have greatly reduced the cost of accessing crystal structures, calculated properties, and experimental records. However, their primary roles are storage, sharing, retrieval, and high-throughput data aggregation~\citep{wang2023secured}. 
Automated quality assessment for scientific reliability remains comparatively limited. Data governance for materials machine learning must simultaneously handle statistical outliers, semantic mismatches, unreasonable structures, failed calculations, and violations of domain rules. 
Therefore, materials data quality assessment should evaluate not only statistical abnormality, but also scientific plausibility and suitability for downstream learning tasks~\citep{chang2022towards}.

\subsection{Scientific data quality assessment and anomaly identification}

Although anomaly detection provides an effective mechanism for identifying suspicious records, conventional approaches usually define anomalies according to statistical rarity rather than scientific reliability. 
Classical anomaly detection approaches include distance-based, density-based~\citep{aydin2023boundary}, statistical, clustering-based~\citep{sanchez2025comparative}, and one-class classification methods. 
In materials data, anomaly detection methods generally target three aspects. The first concerns crystal-structure plausibility, including interatomic distances, lattice parameters, space groups, and coordination environments. The second focuses on property reliability, including formation energy, band gap, energy above hull, and calculation convergence. The third addresses multi-source data consistency, including missing values, duplicate entries, field mismatches, unit inconsistency, and abnormal property relationships.

The main limitation of purely data-driven anomaly detection is that it measures deviation from the current data distribution, but cannot directly determine whether a sample obeys materials-science rules~\citep{liu2025domain}. For example, negative band gaps, out-of-range space-group identifiers, and inconsistencies between lattice matrices and lattice constants should be treated as low-quality evidence even when they are not extreme in a low-dimensional statistical projection. 
These limitations indicate that reliable materials data assessment requires a unified framework that integrates statistical evidence, domain constraints, and relational information among data entities~\citep{feng2023unsupervised}.

\subsection{Domain knowledge integration}
Domain knowledge integration aims to convert expert rules, physical constraints, and empirical relationships into explicit quality indicators. Knowledge-assisted approaches have shown that materials rules can improve quality evaluation and provide explanations that domain experts can audit. Representative methods, such as DKA-DAD~\citep{liu2025domain} and DADMmdk~\citep{SHISiqi_1311}, introduce materials rules into anomaly detection and demonstrate the value of expert knowledge in identifying scientifically inconsistent records.

However, two limitations remain. First, rules are often triggered in isolation, rather than fused with statistical evidence in a unified scoring mechanism. Second, rule evidence is difficult to propagate between samples and features, so local defects may be handled as isolated events instead of being incorporated into a globally consistent quality judgment. 
These limitations motivate the development of methods that can integrate rule-based evidence, statistical signals, and relational structures into a unified quality representation.

\subsection{Graph learning and confidence quantification}
Graph-based methods provide a natural way to capture dependencies among samples, features, and reference data in structured datasets~\citep{zhu2026graph}. Graph propagation approaches enable confidence information to spread through relational structures, while graph attention mechanisms learn adaptive importance among neighboring entities.
Personalized PageRank introduces restart-based propagation and is suitable for spreading confidence over a sample--feature bipartite graph~\citep{yang2024efficient}. Graph attention networks~\citep{vrahatis2024graph} allow each node to assign adaptive weights to its neighbors and can suppress the influence of abnormal neighbors during representation learning. In a data-quality setting, graph propagation provides global consistency, while graph attention provides local anomaly isolation~\citep{ma2024graph}.

Because data quality is influenced by interactions among samples, features, and historical references, graph-based methods provide a promising representation for modeling quality propagation.
Confidence-based learning extends discrete error identification to probabilistic quality estimation~\citep{guo2026sample}. This idea provides useful inspiration for modeling data quality as continuous confidence rather than binary normal/anomalous labels. 
This perspective motivates extending confidence modeling from individual prediction reliability to structured scientific data quality assessment. 
Moreover, historical quality anchors provide a reference for batch-level comparison. The Memory-GAT module uses these anchors to relate current batch scores to a common historical reference.

Overall, existing studies have advanced materials data quality control from statistical screening toward knowledge-assisted evaluation. However, a unified framework that simultaneously considers domain constraints, statistical evidence, relational structures, and historical-reference batch comparison remains insufficiently explored. Addressing this gap requires a continuous and interpretable quality representation capable of integrating heterogeneous quality signals across materials datasets.

\section{Method}
The framework assigns confidence scores to samples and features, then updates them using statistical checks, materials knowledge and graph relationships. These scores express data reliability on a scale from zero to one. They are used to rank records and identify fields that need inspection; they are not estimates of prediction uncertainty. The workflow consists of prior construction, evidence calibration, sample--feature propagation, local attention refinement and historical-reference scoring (Fig.~\ref{fig:framework}).

\rev{Let the dataset be $X=\{x_i\}_{i=1}^{n}$, where $x_i\in\mathbb{R}^{d}$ denotes the $i$-th sample, and $n$ and $d$ denote the numbers of samples and features, respectively. Sample-confidence scores at successive stages are denoted by subscripts: $C_{\mathrm{init}}$ for initial confidence, $C_{\mathrm{cal}}$ for calibrated confidence, $C_{\mathrm{graph}}$ for graph-propagated confidence, and $C_{\mathrm{final}}$ for final confidence. Feature confidence is denoted analogously by $Z_{\mathrm{init}}$, $Z_{\mathrm{cal}}$, and $Z_{\mathrm{graph}}$. A lower confidence score indicates a higher likelihood of quality defects.}

\subsection{Nine-dimensional quality prior}
The initial feature confidence combines nine indicators derived from metadata and numerical data. The indicators translate broad data-quality concepts~\citep{alexakis2025evaluating} into quantities that can be calculated for each field. Let $q_{j,k}$ be the score of feature $j$ on indicator $k$, with weight $v_k$. The initial feature and sample scores are
\begin{equation}
Z_{\mathrm{init},j}=\sum_{k=1}^{9}v_kq_{j,k},\qquad
C_{\mathrm{init},i}=\begin{cases}0.2,&\text{exact duplicate numerical row},\\1,&\text{otherwise}.\end{cases}
\label{eq:init}
\end{equation}
The feature weights sum to one and are fixed at $(0.05,0.15,0.15,0.08,0.20,0.05,0.10,0.10,0.12)$ for documentation availability, statistical-property matching, numeric compatibility, metadata coupling, non-missing rate, uniqueness, distributional skewness, numerical regularity and descriptor-variation relevance, respectively. Every record in an exact duplicate group receives the lower initial sample score. The weighted feature indicators and the sample scores are then updated by the evidence checks.

The weights are fixed throughout the main experiments. Appendix~\ref{app:prior_details} defines each indicator and explains which become constant after preprocessing. The last indicator measures the correlation between a feature and the first principal component of the input descriptors. Section~\ref{sec:robustness} examines sensitivity to the weights and penalty coefficients.

\inlinefig{fig:framework}{0.92\linewidth}{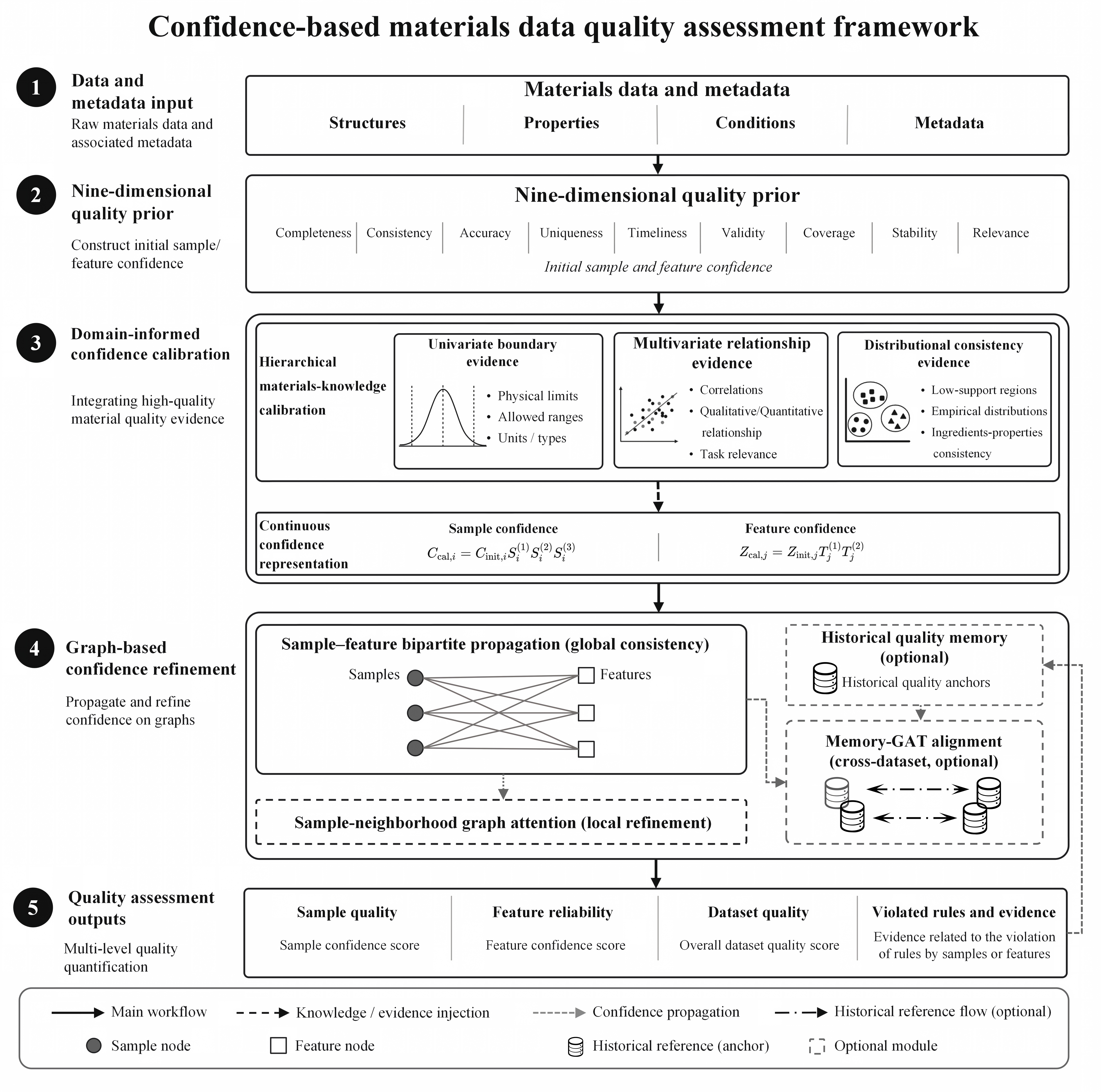}{Workflow of the proposed materials data quality assessment framework. The diagram summarizes prior construction, evidence calibration, graph refinement and quality outputs. The feature indicators are defined in Section~3.1. The schematic calibration factors are implemented using the minimum cell penalty for each sample and the mean cell penalty for each feature, as specified in Eq.~\ref{eq:cal}.}

\inlinetable{tab:notation}{Main notation used in the confidence-assessment workflow.}{%
\small
\begin{adjustbox}{max width=\linewidth}
\begin{tabular}{l p{.72\linewidth}}
\toprule
Symbol & Meaning \\
\midrule
$q_{j,k},v_k$ & Feature indicator $k$ and its normalized weight \\
$C_{\mathrm{init}},Z_{\mathrm{init}}$ & Initial sample confidence accounting for duplicates, and weighted feature prior \\
$C_{\mathrm{cal}},Z_{\mathrm{cal}}$ & Sample and feature confidence after evidence calibration \\
$a_{ij},B_{ij}$ & Cell penalty and sample--feature edge reliability \\
$P,\alpha$ & Row-stochastic transition and restart coefficient \\
$C_{\mathrm{graph}},Z_{\mathrm{graph}}$ & Scores after bipartite propagation \\
$r_i,C_{\mathrm{final},i}$ & Rule-trained GAT risk and final sample confidence \\
$m_i,Q_D$ & Historical-reference sample score and batch mean \\
\bottomrule
\end{tabular}
\end{adjustbox}
}

\subsection{Hierarchical materials-knowledge calibration}
A record can fall within a common numerical range while violating a materials constraint. Calibration therefore uses three types of evidence: individual field ranges, relationships between fields and deviations from the batch distribution~\citep{WEIXiao:386}. In the controlled benchmark, these checks cover configured value ranges, agreement between lattice representations, and standardized-score/IQR checks. Property targets, identifiers and injected anomaly labels are excluded from scoring.

Evidence is mapped to a cell penalty $a_{ij}$: severe evidence gives 0.3, mild evidence gives 0.7, and unaffected cells give 1.0. Severe evidence takes precedence, so overlapping checks do not multiply the same cell repeatedly. The calibrated confidence scores are
\begin{equation}
C_{\mathrm{cal},i}=C_{\mathrm{init},i}\min_j a_{ij},\qquad
Z_{\mathrm{cal},j}=Z_{\mathrm{init},j}\frac1n\sum_i a_{ij}.
\label{eq:cal}
\end{equation}
Here $S_i=\min_j a_{ij}$ and $T_j=n^{-1}\sum_i a_{ij}$ are the sample and feature factors used in the workflow. The evidence-channel factors shown schematically in Fig.~\ref{fig:framework} are implemented by these aggregate penalties: the minimum cell penalty for a sample and the mean cell penalty for a feature. The severe and mild coefficients remain fixed in the main experiments and are varied in the sensitivity analysis.

The sample-feature reliability matrix uses $B_{ij}=1$, 0.5 and 0.1 for unaffected, mildly flagged and severely flagged cells. Table~\ref{tab:rules} specifies the implemented checks and their screening ranges.
\inlinetable{tab:rules}{Representative quality-evidence types and their effects on confidence.}{%
\small
\begin{adjustbox}{max width=\linewidth}
\begin{tabular}{p{.18\linewidth}p{.48\linewidth}p{.25\linewidth}}
\toprule
Evidence & Evidence content & Interpretation \\
\midrule
Univariate boundary & Band gap [0,10] eV; energy above hull [0,0.5] eV/atom; space group [1,230]; lattice lengths [2,30] \AA; angles [30,150] degrees; fractional coordinates [0,1]. Outside a range is severe. & Configured plausibility checks; coordinate wrapping must be declared. \\
Multivariate relationship & Relative mismatch between a lattice length and its lattice-matrix row norm exceeds 0.05; all four involved cells receive severe evidence. & Inconsistency between stored representations. \\
Global distribution & Population-standardized $|z|>5$: severe; $|z|>3$ or outside the Tukey $1.5\,\mathrm{IQR}$ interval: mild unless severe. Zero standard deviation or IQR is replaced by one. & Batch-level deviation evidence. \\
\bottomrule
\end{tabular}
\end{adjustbox}
}

\subsection{Sample-feature bipartite graph propagation}

Calibration first scores each record and field from the checks they trigger. A bipartite graph then links sample nodes to feature nodes~\citep{wei2025graph}, using $B_{ij}$ as the weight between sample $i$ and feature $j$. Each sample receives information from its associated features, and each feature receives information from the samples containing it.

This exchange allows the confidence of a record to reflect both its own checks and the reliability of its fields. In turn, feature confidence reflects the records in which that feature appears~\citep{zhang2025your}. Let
\[
H_{\mathrm{cal}}=
\begin{bmatrix}
C_{\mathrm{cal}} \\
Z_{\mathrm{cal}}
\end{bmatrix}
\]
collect the calibrated sample and feature scores. The propagation matrix is
\begin{equation}
P=\begin{bmatrix}
0 & D_s^{-1}B \\
D_f^{-1}B^{\top} & 0
\end{bmatrix},
\label{eq:block}
\end{equation}
where $D_s=\operatorname{diag}(B\mathbf1_d)$ and $D_f=\operatorname{diag}(B^\top\mathbf1_n)$ are the separate sample and feature degree matrices. Thus each outgoing block is row-normalized. The confidence propagation process is then written as
\begin{equation}
H^{(t+1)}=\alpha H_{\mathrm{cal}}+(1-\alpha)PH^{(t)}.
\label{eq:prop}
\end{equation}
Here, $\alpha=0.35$ controls how much calibrated confidence is retained at each iteration~\citep{liao2023efficient}. The first term preserves the original evidence, and the second averages the scores of neighboring nodes using the weights in $P$. Starting from $H^{(0)}=H_{\mathrm{cal}}$, iteration stops when the largest score change is below $10^{-8}$, or after 200 iterations. The resulting sample and feature scores are $C_{\mathrm{graph}}$ and $Z_{\mathrm{graph}}$. Appendix~\ref{app:convergence} establishes convergence and checks the implementation against independent calculations.

\subsection{Graph attention local refinement}
Bipartite propagation combines information across the sample--feature graph. Local refinement then uses a sample KNN graph to compare records with similar descriptors. Attention learns how strongly each neighboring record contributes to the risk score.

For each sample, the normalized raw feature vector is first weighted by graph-propagated feature confidence and then concatenated with the sample's own graph-propagated confidence:
\begin{equation}
U_i=\left[\tilde{x}_i\odot Z_{\mathrm{graph}} \,\Vert\, C_{\mathrm{graph},i}\right],
\label{eq:gatinput}
\end{equation}
where $\tilde{x}_i$ denotes the normalized feature vector, $\odot$ denotes elementwise multiplication, and $\Vert$ denotes vector concatenation. This operation reduces the influence of low-confidence features and preserves more information from high-confidence features. A sample KNN graph is then constructed from $U_i$, and graph attention learns adaptive neighbor weights~\citep{kaibiao2023adaptive}. 
The KNN graph has eight incoming neighbors and a self-loop per sample. The two-layer GAT uses four first-layer heads with 16 outputs each, ELU and dropout 0.05, followed by a sigmoid risk output. It is trained for 50 full-batch epochs with Adam (learning rate $10^{-3}$, weight decay $10^{-5}$, gradient clipping one). The training target is the rule-derived risk $t_i=1-C_{\mathrm{cal},i}$:
\begin{equation}
\mathcal L_{\mathrm{GAT}}=\frac1n\sum_i(r_i-t_i)^2.
\label{eq:gatloss}
\end{equation}
Neither injected anomaly labels nor downstream property targets enter this loss. The model learns to reproduce this rule-derived risk using the sample and its neighbors.
Let $r_i\in[0,1]$ be the risk score output by the GAT, where a larger value indicates a more suspicious sample. The final sample confidence is updated as
\begin{equation}
C_{\mathrm{final},i}
=
\mathrm{clip}\left(C_{\mathrm{graph},i}(1-\lambda r_i),0,1\right),
\label{eq:final}
\end{equation}
where $\lambda=0.5$ controls the impact of GAT refinement. If the GAT estimates high risk, the confidence is further decreased from the graph-propagated value. If the risk is low, the original confidence is largely retained. A matched GAT configuration uses $C_{\mathrm{cal}},Z_{\mathrm{cal}}$ in place of the propagated scores, with the same objective and training budget. These comparisons distinguish the contributions of local attention and bipartite propagation.

\subsection{Historical-reference Memory-GAT}
Traditional anomaly scores are dataset-dependent. Historical quality memory provides a reference for comparing newly arrived batches with previously assessed samples. The memory stores descriptors, confidence, rule evidence and sample fingerprints. Historical samples covering a range of confidence scores are selected as frozen anchors~\citep{VO202296}. Their reference scores are obtained through rule calibration and graph propagation. The anchor descriptors and scores remain fixed while the attention model is fitted for the current batch.

The memory graph uses standardized descriptors and transformed reference confidence scores. Each current node receives messages from eight historical neighbors, six current neighbors and itself. Historical nodes receive only their own self-loop messages. A two-layer attention encoder with a sigmoid output is trained to reproduce the historical reference scores and the current rule/propagation scores. A smoothness term encourages connected nodes to have similar outputs. Appendix~\ref{app:memory_details} gives the inputs, loss and training settings. Injected anomaly labels are excluded from training.

For current-sample scores $m_i$, the batch score is
\begin{equation}
Q_D=\frac1{|D|}\sum_{i\in D}m_i.
\label{eq:datasetquality}
\end{equation}
The mean $Q_D$ summarizes the quality of the current batch relative to the historical anchors. Section~5.5 compares its ordering of controlled batches with batch-only scores and a distance-weighted historical KNN reference.

\subsection{Quality representation and applications}

The framework outputs sample confidence
$C_{\mathrm{final},i}$, feature confidence
$Z_{\mathrm{graph},j}$, and the corresponding rule-trigger
evidence. Sample confidence supports quality ranking and
helps users prioritize low-confidence records for inspection.
Feature confidence summarizes the quality of individual
fields and helps identify columns requiring further checks.
Rule-trigger evidence specifies the affected fields and
the checks triggered by each record, providing a basis
for verification against the original data source.

When historical quality memory is available, the framework
uses historical anchors to calculate a reference-based score
$m_i$ for each current sample. The dataset-level quality score
$Q_D$ is the mean of these scores over the current batch,
supporting comparison of overall quality across batches.
Appendix~\ref{app:workflow} presents the complete assessment
workflow.

\section{Experimental setup}
The experiments assess anomaly identification, materials data screening, the contribution of each module, interpretation of flagged records and comparison of batches. They include 23 ODDS datasets~\citep{tsai2025anollm}, controlled contamination of mp\_20 with downstream prediction, ablation and sensitivity tests, an alloy-data case study and historical-reference batch scoring.

\subsection{Generic anomaly identification benchmark}

The ODDS experiment tests anomaly identification without materials-specific rules. It retains the feature prior, statistical checks and graph refinement.

The method is evaluated on 23 ODDS datasets using AUC-ROC. The suite includes datasets that could be retrieved and processed consistently. Datasets with unavailable download links and exceptionally large datasets with very few features were excluded.

The baselines cover statistical methods, neighborhood-based detectors, representation-learning approaches, and large-language-model-based anomaly detection~\citep{tsai2025anollm}.

\subsection{Materials data cleaning and downstream validation}
The mp\_20 dataset contains 45,229 records. Five independently sampled 16,000-record subsets (seeds 0--4) are deduplicated by material identifier and grouped by exact composition label and nine unrounded lattice-matrix entries. Grouped splits yield approximately 12,000 outer-training and 4,000 clean test records; the outer-training partition is further divided into 10,000 fitting and 2,000 calibration records. Feature eligibility and median imputation are determined from outer training, excluding identifiers, property targets and auxiliary property fields. The main runs retain 43 numerical descriptors.

Corruption is applied only after splitting. Half of each fitting or calibration batch is contaminated, evenly across statistical outliers, configured boundary violations, lattice-representation inconsistency and mixed perturbations. The same records receive zero-mean target noise with standard deviation 2.8 times the clean batch target standard deviation. The mixed-input perturbation scale is $0.6\sigma$. This setup tests screening when both descriptors and target values contain errors. Appendix~\ref{app:corruption} gives the corruption procedure.

Scoring uses input descriptors and the evidence derived from them. The formation-energy target is excluded from prior construction, scoring, graph construction and GAT training. For local GAT refinement, fitting and calibration batches are each standardized using their own column means and standard deviations; a local GAT is trained separately on each batch using its rule-derived risk targets. For each method, the threshold that maximizes F1 on the synthetically labeled calibration batch is then applied to the independently scored fitting batch. The clean test data are used only for downstream evaluation.

HistGradientBoostingRegressor~\citep{raparthi2024implementation} uses 220 iterations, learning rate 0.055 and L2 regularization 0.02. All methods share the same clean test records within each seed. Isolation Forest (50 estimators, contamination 0.1), PCA (95\% variance) and random retention provide comparisons. For each seed, the random control retains the same number of fitting records as the corresponding screening method. In the tables, the labels rules and full identify the method whose retained count is matched; random selection itself uses no confidence scores.

MACE~\citep{kovacs2023evaluation} is evaluated on five grouped 4,000-record subsets with approximately 2,500 fitting, 500 clean validation and 1,000 clean test crystals before shared structural eligibility filtering. Thresholds use separately corrupted copies of the calibration partition. Every method shares the eligible clean test set; the clean validation loss selects the model, and test predictions are computed once after training. The repeated subset study comprises five splits and seven conditions, totaling 35 MACE training runs. MACE uses random initialization, two interactions, hidden irreps \texttt{128x0e+128x1o}, radius 4~\AA, float32, batch size 32, learning rate 0.005 and a common ten-epoch budget. Only energy is trained; forces have zero loss weight and stage two is disabled. Appendix~\ref{app:mace_details} specifies structural filtering, remaining settings and evaluation procedure.

\subsection{Ablation and robustness analysis}
The ablation compares six configurations: statistical checks, boundary checks, complete rules, rules with propagation, rules with GAT, and rules with both graph stages. This separates the effects of relationship checks, sample--feature propagation and local attention. Historical memory is tested in the batch-comparison experiment.

Robustness crosses contamination fractions 10\%, 30\% and 50\% with mixed-descriptor scales $0.1\sigma$, $0.3\sigma$ and $0.6\sigma$, while holding other corruption mechanisms and target-noise multipliers fixed. Parameter sensitivity includes nine severe/mild penalty combinations, equal feature weights, and individual weight perturbation or omission. All configurations are repeated over seeds 0--4; the full weight results are in Appendix~\ref{app:sensitivity}. Additional diagnostics use a fixed threshold or a prespecified 50\% deletion budget; these are distinguished from the main results, which use thresholds selected on the calibration partition.

\subsection{Materials data governance case study}

To demonstrate interpretability in a realistic materials data governance scenario, this study uses an alloy mechanical-property dataset collected from public literature~\citep{xu2025intelligent,wang2023rapid,chen2023introduction}.
The primary case-study table contains 185 records and 50 fields, comprising one alloy identifier, 14 elemental-composition variables, one testing-temperature variable, two mechanical-property variables---ultimate tensile strength (UTS) and fracture elongation (FE)---and 32 composition-derived physicochemical, thermodynamic, and elastic descriptors. A larger source-literature reference table containing 315 records and 61 fields is used as an auxiliary source for statistical comparison and fitting-based verification. It comprises one sample identifier, the same 14 elemental-composition variables, 12 processing-history variables describing homogenization, deformation, rolling, annealing, and aging, the two mechanical-property variables, and the same 32 material descriptors. Detailed definitions and categories of these fields are provided in Appendix~\ref{app:alloy_fields}.

The case study examines whether each flagged record can be linked to a specific quality concern. It combines range checks, relationships between fields, distribution checks, missingness and differences between reported and model-predicted properties. The audit also records whether processing-history information is available.

Samples outside configured screening ranges or with property differences at matched recorded composition and temperature are treated as review candidates. Processing-history availability is included in the review evidence.
Samples identified primarily through statistical or global distribution evidence are further examined using KNN and Random Forest regressors trained under five-fold cross-validation~\citep{dinesh2024medical}.
The regressors predict ultimate tensile strength (UTS) and fracture elongation (FE), and large fitting residuals are used as additional evidence of potential inconsistency~\citep{li2022relationship}.
The goal is to evaluate whether the proposed method can generate auditable and interpretable quality assessments for public materials datasets.

\subsection{Historical-reference batch comparison}
Mutually disjoint mp\_20 samples form three historical batches of 800 records at 0\%, 10\% and 30\% contamination. Historical samples are sorted by their rule/propagation confidence scores, and 800 anchors are selected at evenly spaced positions in this ordering. Four current batches contain 900 records each at 0\%, 10\%, 30\% and 50\% contamination. All preprocessing for the 18 shared descriptors is fitted on history only.

The experiment is repeated over five seeds and compares batch-only scores, distance-weighted historical KNN and Memory-GAT. Mean scores and Spearman correlation with decreasing contamination assess controlled batch ordering. The experiment evaluates historical reference scoring across disjoint batches from a common materials source. Current contamination labels are used for evaluation.

\subsection{Evaluation metrics and parameter settings}

Anomaly ranking is evaluated by AUC-ROC and average precision (AP). Screening is evaluated by F1 at the calibration-selected threshold, with recall also reported for each corruption mechanism. Downstream prediction is evaluated by the coefficient of determination ($R^2$) and mean absolute error (MAE). Higher AUC-ROC, AP, F1, recall and $R^2$, and lower MAE, indicate better performance. For batch comparison, Spearman correlation measures agreement between batch confidence and decreasing contamination, so a positive correlation indicates higher scores for cleaner batches.

The ODDS benchmark retains its original seed-42 results. The materials experiments are repeated with seeds 0--4 and report mean $\pm$ sample standard deviation. Within each seed, methods share the data split and corruption realization. Differences in $R^2$ are calculated within each seed and summarized with 95\% Student-$t$ intervals over the five runs. Default severe/mild penalties are 0.3/0.7, edge weights are 0.1/0.5/1.0, restart is 0.35, and the GAT gate is 0.5. These manual settings remain fixed except in declared sensitivity experiments. KNN size is eight, the local GAT has two layers and hidden width 64, and historical memory retains 800 anchors. Full training and runtime settings are reported in the method and appendix.

\section{Results and discussion}

\subsection{Generalization evaluation on ODDS benchmarks}
Table~\ref{tab:odds_summary} summarizes the average AUC-ROC and overall ranking on the ODDS benchmarks. Detailed per-dataset results are provided in Appendix~\ref{app:odds}. 
With all materials-domain rules disabled, the proposed method achieves an average AUC-ROC of 0.866 on 23 ODDS datasets, showing that the generic configuration can rank anomalies without materials-specific rules.

\inlinetable{tab:odds_summary}{Average AUC-ROC and ranking on ODDS benchmarks.}{%
\centering
\resizebox{0.45\linewidth}{!}{%
\begin{tabular}{lcc}
\toprule
Method & Average AUC-ROC & Rank \\
\midrule
AnoLLM-135M & 0.892 & 1 \\
ICL & 0.884 & 2 \\
DTE & 0.882 & 3 \\
KNN & 0.878 & 4 \\
AnoLLM-360M & 0.867 & 5 \\
\textbf{Our method} & \textbf{0.866} & \textbf{6} \\
NeuTraL & 0.863 & 7 \\
SLAD & 0.848 & 8 \\
RCA & 0.843 & 9 \\
IForest & 0.839 & 10 \\
PCA & 0.814 & 11 \\
DeepSVDD & 0.812 & 12 \\
ECOD & 0.781 & 13 \\
REPEN & 0.756 & 14 \\
GOAD & 0.729 & 15 \\
\bottomrule
\end{tabular}
}

\tablenote{AUC-ROC is the area under the receiver operating characteristic curve; the average is over 23 datasets. Rank orders methods by decreasing average AUC-ROC (1 is highest).}
}
The ODDS results establish the anomaly-ranking performance of the generic configuration. The following materials experiments examine the contributions of rule calibration and graph refinement.

\subsection{Materials data quality assessment and downstream model validation}
The downstream experiments test whether screening improves models trained on contaminated materials data. Table~\ref{tab:mace} reports MACE results, and Table~\ref{tab:ablation} compares screening methods and graph modules using HGBR. All methods are evaluated on the same clean test records within each seed.
\inlinetable{tab:mace}{MACE downstream validation on five repeated grouped subsets (mean $\pm$ SD). F1 and retained counts follow the common structural eligibility check. MAE is in eV/atom. Matched random retains the same number of structures as complete rules.}{%
\small
\begin{adjustbox}{max width=\linewidth}
\begin{tabular}{lcccc}
\toprule
Condition & F1 & $R^2$ & MAE & Retained \\
\midrule
Clean reference & -- & 0.9387 $\pm$ 0.0158 & 0.1488 $\pm$ 0.0195 & 2256 $\pm$ 22 \\
No cleaning & 0.0000 $\pm$ 0.0000 & 0.5896 $\pm$ 0.0527 & 0.4908 $\pm$ 0.0248 & 2256 $\pm$ 22 \\
Complete rules & 0.9622 $\pm$ 0.0039 & 0.8801 $\pm$ 0.0268 & 0.2121 $\pm$ 0.0316 & 1158 $\pm$ 8 \\
Rules + propagation + GAT & 0.9451 $\pm$ 0.0119 & 0.8150 $\pm$ 0.0413 & 0.2790 $\pm$ 0.0328 & 1185 $\pm$ 49 \\
Isolation Forest & 0.7002 $\pm$ 0.0916 & 0.6589 $\pm$ 0.0602 & 0.4335 $\pm$ 0.0257 & 1350 $\pm$ 394 \\
PCA & 0.6429 $\pm$ 0.0256 & 0.5740 $\pm$ 0.0576 & 0.4963 $\pm$ 0.0304 & 1466 $\pm$ 146 \\
Matched random & 0.4761 $\pm$ 0.0124 & 0.2720 $\pm$ 0.1468 & 0.6444 $\pm$ 0.0541 & 1158 $\pm$ 8 \\
\bottomrule
\end{tabular}
\end{adjustbox}

\tablenote{F1 is the harmonic mean of anomaly-detection precision and recall. $R^2$ is the coefficient of determination on clean test structures; MAE is the mean absolute prediction error in eV/atom. Retained is the number of fitting structures kept after screening and the shared structural check. Values are mean $\pm$ sample standard deviation over five seeds; -- indicates a metric not reported for that condition.}
}

On the shared eligible fitting set of $2256\pm22$ structures, clean-reference MACE achieves $R^2=0.9387\pm0.0158$, compared with $0.5896\pm0.0527$ without cleaning. Rule calibration retains $1158\pm8$ structures and recovers $R^2$ to $0.8801\pm0.0268$. Isolation Forest and PCA obtain $0.6589\pm0.0602$ and $0.5740\pm0.0576$, respectively. The rule-versus-no-cleaning paired gain is 0.2906, with a descriptive 95\% interval [0.2098,0.3714]. The full graph configuration yields $0.8150\pm0.0413$; its difference from rules is $-0.0652$ [\(-0.1321,0.0018\)]. Rule-based screening therefore recovers a substantial part of the prediction performance lost to corruption under the shared ten-epoch budget. Test sets contain 990--994 eligible clean crystals per seed.
\inlinetable{tab:ablation}{Independent module ablation under 50\% mixed contamination (five-seed mean $\pm$ SD). Retention is out of approximately 10,000 fitting records; MAE is in eV/atom. Random controls match the retained counts of complete rules (rules) or the configuration with both graph stages (full).}{%
\small
\begin{adjustbox}{max width=\linewidth}
\begin{tabular}{lccccc}
\toprule
Configuration & AP & F1 & $R^2$ & MAE & Retained \\
\midrule
Clean reference & -- & -- & 0.6566 $\pm$ 0.0088 & 0.4222 $\pm$ 0.0105 & 10000 $\pm$ 0 \\
No cleaning & -- & -- & 0.5379 $\pm$ 0.0077 & 0.5154 $\pm$ 0.0110 & 10000 $\pm$ 0 \\
Statistical & 0.8010 $\pm$ 0.0089 & 0.8687 $\pm$ 0.0064 & 0.5682 $\pm$ 0.0155 & 0.4791 $\pm$ 0.0149 & 3638 $\pm$ 85 \\
Boundary & 0.8534 $\pm$ 0.0073 & 0.8695 $\pm$ 0.0064 & 0.5785 $\pm$ 0.0178 & 0.4714 $\pm$ 0.0149 & 3563 $\pm$ 81 \\
Complete rules & 0.9427 $\pm$ 0.0037 & 0.9705 $\pm$ 0.0019 & 0.6160 $\pm$ 0.0140 & 0.4495 $\pm$ 0.0132 & 4696 $\pm$ 21 \\
Rules + propagation & 0.9024 $\pm$ 0.0054 & 0.7715 $\pm$ 0.2776 & 0.5828 $\pm$ 0.0433 & 0.4760 $\pm$ 0.0313 & 6039 $\pm$ 1817 \\
Rules + GAT & 0.9516 $\pm$ 0.0039 & 0.9653 $\pm$ 0.0055 & 0.6137 $\pm$ 0.0145 & 0.4508 $\pm$ 0.0136 & 4755 $\pm$ 84 \\
Rules + propagation + GAT & 0.9530 $\pm$ 0.0041 & 0.9634 $\pm$ 0.0017 & 0.6107 $\pm$ 0.0143 & 0.4521 $\pm$ 0.0135 & 4707 $\pm$ 72 \\
Isolation Forest & 0.8687 $\pm$ 0.0233 & 0.7731 $\pm$ 0.0194 & 0.5403 $\pm$ 0.0198 & 0.5028 $\pm$ 0.0180 & 5264 $\pm$ 917 \\
PCA & 0.8144 $\pm$ 0.0042 & 0.7177 $\pm$ 0.0134 & 0.5067 $\pm$ 0.0183 & 0.5260 $\pm$ 0.0141 & 4748 $\pm$ 559 \\
Matched random (rules) & -- & -- & 0.4773 $\pm$ 0.0122 & 0.5490 $\pm$ 0.0133 & 4696 $\pm$ 21 \\
Matched random (full) & -- & -- & 0.4749 $\pm$ 0.0136 & 0.5492 $\pm$ 0.0113 & 4707 $\pm$ 72 \\
\bottomrule
\end{tabular}
\end{adjustbox}

\tablenote{AP is average precision for anomaly ranking; F1 measures screening at the calibration-selected threshold. $R^2$ and MAE describe HGBR predictions on the clean test set. Retained counts fitting records kept for training. Values are mean $\pm$ sample standard deviation over five seeds; -- indicates a metric not reported for that condition.}
}

The HGBR results show that contamination reduces clean-reference $R^2$ from $0.6566\pm0.0088$ to $0.5379\pm0.0077$. Statistical and boundary calibration recover part of the loss, while complete rules achieve $0.6160\pm0.0140$. The paired gain over no cleaning is 0.0781 [0.0683,0.0880]. Matched-count random retention yields markedly lower $R^2$, showing that evidence-based sample selection drives the recovery.

The full graph configuration achieves the highest AP, increasing it from $0.9427\pm0.0037$ for complete rules to $0.9530\pm0.0041$. Among the screening configurations, complete rules achieve the highest F1 ($0.9705\pm0.0019$) and downstream $R^2$ ($0.6160\pm0.0140$), compared with $0.9634\pm0.0017$ and $0.6107\pm0.0143$ for the full graph configuration. The paired $R^2$ difference is $-0.0053$ [\(-0.0121,0.0016\)]. Propagation alone gives AP $0.9024\pm0.0054$ and more variable retained counts. Local attention therefore improves anomaly ranking in this benchmark, while rule calibration gives the highest mean F1 and test $R^2$ among the screening configurations.

\inlinefig{fig:ablation}{\linewidth}{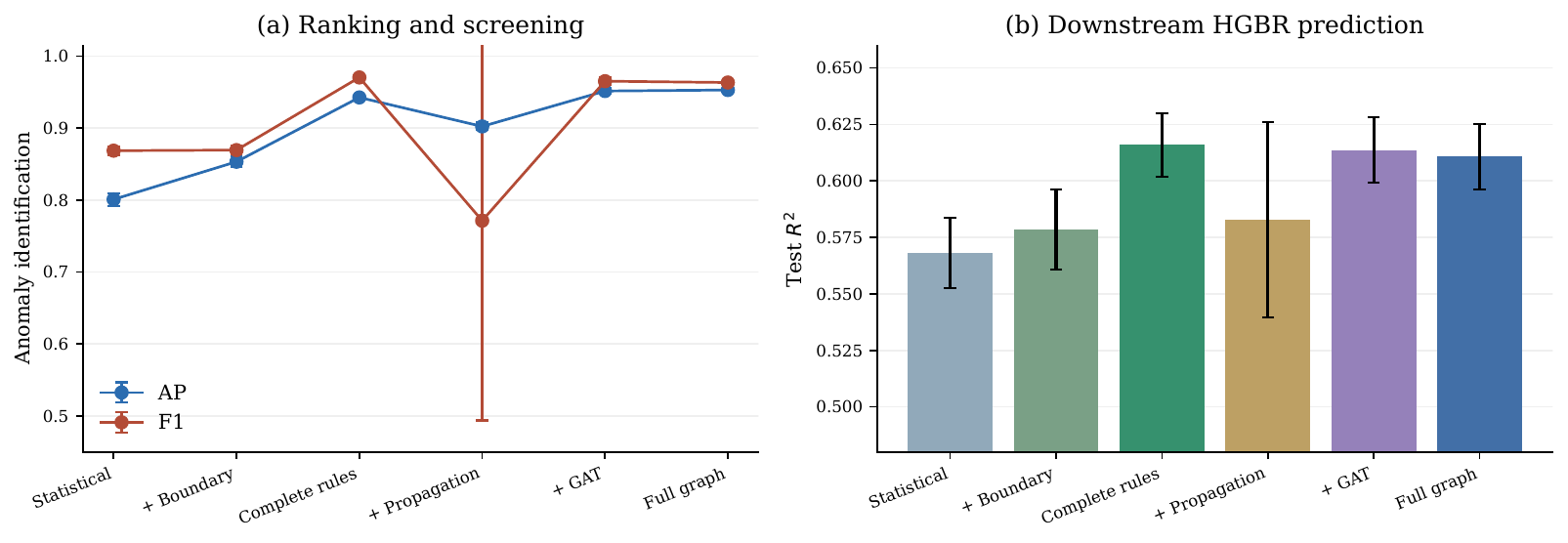}{Repeated ablation under mixed contamination. Rules, bipartite propagation and local GAT are evaluated separately for anomaly ranking and downstream prediction.}

\subsection{Contamination robustness and parameter sensitivity}\label{sec:robustness}
Table~\ref{tab:noise} compares screening F1 at three contamination fractions and three mixed-input perturbation scales. Only the mixed-descriptor perturbation changes; the other corruption mechanisms and target noise remain fixed. F1 is higher at larger contamination fractions for both configurations and changes little across the tested perturbation scales.
\inlinetable{tab:noise}{Screening F1 across contamination fractions and mixed-descriptor scales (five-seed mean $\pm$ SD).}{%
\small
\begin{adjustbox}{max width=\linewidth}
\begin{tabular}{llccc}
\toprule
Configuration & Contamination & $0.1\sigma$ & $0.3\sigma$ & $0.6\sigma$ \\
\midrule
Rules & 10\% & 0.782 $\pm$ 0.010 & 0.782 $\pm$ 0.010 & 0.782 $\pm$ 0.010 \\
Rules & 30\% & 0.931 $\pm$ 0.002 & 0.931 $\pm$ 0.002 & 0.931 $\pm$ 0.002 \\
Rules & 50\% & 0.970 $\pm$ 0.002 & 0.971 $\pm$ 0.002 & 0.970 $\pm$ 0.002 \\
Full graph & 10\% & 0.733 $\pm$ 0.016 & 0.728 $\pm$ 0.021 & 0.728 $\pm$ 0.021 \\
Full graph & 30\% & 0.905 $\pm$ 0.030 & 0.911 $\pm$ 0.018 & 0.910 $\pm$ 0.019 \\
Full graph & 50\% & 0.962 $\pm$ 0.001 & 0.963 $\pm$ 0.002 & 0.963 $\pm$ 0.002 \\
\bottomrule
\end{tabular}
\end{adjustbox}

\tablenote{Contamination is the fraction of records with injected errors. Column headings give the mixed-descriptor perturbation scale relative to the descriptor standard deviation $\sigma$. Entries are F1 scores (higher is better), reported as mean $\pm$ sample standard deviation over five seeds. Full graph uses both propagation and GAT.}
}

Table~\ref{tab:mechanism} reports recall separately for the four corruption mechanisms. Complete rules detect almost all injected records at the calibration-selected thresholds. The F1 results in Table~\ref{tab:ablation} also account for clean records incorrectly removed during screening.
\inlinetable{tab:mechanism}{Recall by injected mechanism under internal calibration (mean $\pm$ SD; approximately 1,250 contaminated fitting records per mechanism per seed).}{%
\small
\begin{adjustbox}{max width=\linewidth}
\begin{tabular}{lcccc}
\toprule
Mechanism & Statistical & Rules & Rules + propagation & Full graph \\
\midrule
Statistical outlier & 0.9992 $\pm$ 0.0006 & 1.0000 $\pm$ 0.0000 & 0.6373 $\pm$ 0.5006 & 0.9685 $\pm$ 0.0268 \\
Boundary violation & 0.9979 $\pm$ 0.0018 & 1.0000 $\pm$ 0.0000 & 0.9832 $\pm$ 0.0371 & 0.9990 $\pm$ 0.0010 \\
Representation mismatch & 0.9803 $\pm$ 0.0032 & 1.0000 $\pm$ 0.0000 & 0.6000 $\pm$ 0.5477 & 0.9995 $\pm$ 0.0007 \\
Mixed perturbation & 0.9706 $\pm$ 0.0068 & 0.9998 $\pm$ 0.0004 & 0.7053 $\pm$ 0.4073 & 0.9997 $\pm$ 0.0004 \\
\bottomrule
\end{tabular}
\end{adjustbox}

\tablenote{Entries are recall: the fraction of injected records of each mechanism identified as anomalous. Statistical uses distribution checks only; Rules uses the complete rule set; Full graph adds propagation and GAT. Values are mean $\pm$ sample standard deviation over five seeds.}
}

Appendix~\ref{app:score_diagnostics} reports the score distributions at each stage. The main comparison uses AP, calibration-selected F1 and downstream prediction to assess ranking and screening performance.

Table~\ref{tab:penalty} varies the severe and mild penalties for the propagation configuration. AP ranges from 0.8991 to 0.9052. F1 and downstream $R^2$ vary more because they also depend on applying the calibration threshold to a separately scored fitting batch. Appendix~\ref{app:sensitivity} reports the corresponding tests with equal feature weights and changes to individual weights.
\inlinetable{tab:penalty}{Penalty sensitivity of the static-propagation configuration with independent threshold calibration (five-seed mean $\pm$ SD).}{%
\small
\begin{adjustbox}{max width=\linewidth}
\begin{tabular}{lccc}
\toprule
Severe/mild & AP & F1 & $R^2$ \\
\midrule
s0.2/m0.6 & 0.9052 $\pm$ 0.0052 & 0.7423 $\pm$ 0.3256 & 0.5794 $\pm$ 0.0491 \\
s0.2/m0.7 & 0.9046 $\pm$ 0.0052 & 0.7619 $\pm$ 0.2766 & 0.5803 $\pm$ 0.0385 \\
s0.2/m0.8 & 0.9038 $\pm$ 0.0053 & 0.7903 $\pm$ 0.2220 & 0.5761 $\pm$ 0.0367 \\
s0.3/m0.6 & 0.9034 $\pm$ 0.0053 & 0.7498 $\pm$ 0.3392 & 0.5855 $\pm$ 0.0438 \\
s0.3/m0.7 & 0.9024 $\pm$ 0.0054 & 0.7715 $\pm$ 0.2776 & 0.5828 $\pm$ 0.0433 \\
s0.3/m0.8 & 0.9014 $\pm$ 0.0055 & 0.7960 $\pm$ 0.2179 & 0.5781 $\pm$ 0.0351 \\
s0.4/m0.6 & 0.9010 $\pm$ 0.0056 & 0.7662 $\pm$ 0.4265 & 0.5939 $\pm$ 0.0388 \\
s0.4/m0.7 & 0.9000 $\pm$ 0.0055 & 0.8318 $\pm$ 0.2797 & 0.5970 $\pm$ 0.0353 \\
s0.4/m0.8 & 0.8991 $\pm$ 0.0056 & 0.8164 $\pm$ 0.2111 & 0.5828 $\pm$ 0.0417 \\
\bottomrule
\end{tabular}
\end{adjustbox}

\tablenote{Severe/mild gives the cell penalty coefficients: for example, s0.2/m0.6 means 0.2 for severe evidence and 0.6 for mild evidence. AP is average precision; F1 measures threshold-based screening; $R^2$ evaluates HGBR prediction on clean test data. Values are mean $\pm$ sample standard deviation over five seeds.}
}

\subsection{Interpretable materials data governance case study}
To enhance interpretability in a realistic materials data governance scenario, this study conducts a case analysis on alloy data collected from public literature~\citep{xu2025intelligent}. A curated structured analysis table supports rule aggregation and consistency checking~\citep{wang2023rapid}~\citep{chen2023introduction}. Record-level auditing identifies missing FE in 68 records and missing UTS in 12 records of the 185-record table. These are field-specific counts and can overlap within the same record. The processing-history fields are absent from this table. Two FE values exceed the configured screening threshold. Separately, 15 composition/temperature group--target combinations meet the property-spread criteria (Appendix~\ref{app:case_audit}); these group-level events are distinct from the record-level missingness counts. Table~\ref{tab:alloy_case} retains representative evidence patterns and distinguishes their interpretation.

\inlinetable{tab:alloy_case}
{Representative review candidates and evidence in the public-literature alloy dataset.}{%
\footnotesize
\setlength{\tabcolsep}{2.5pt}
\renewcommand{\arraystretch}{1.08}

\begin{tabularx}{\linewidth}{
>{\raggedright\arraybackslash}p{0.175\linewidth}
>{\raggedright\arraybackslash}p{0.275\linewidth}
>{\raggedright\arraybackslash}X
>{\raggedright\arraybackslash}p{0.14\linewidth}
}
\toprule
Evidence type & Representative sample & Main evidence & Interpretation \\
\midrule

Unusual recorded value &
Ti$_{67}$Al$_{12}$V$_7$Cr$_7$Nb$_7$
(750~$^\circ$C, FE = 130\%) &
Recorded FE = 130\% triggers the configured value-range check. &
Source/context check required \\

\addlinespace[2pt]

Property spread with missing context &
Co$_{19.60}$Cr$_{19.60}$Fe$_{19.60}$Ni$_{19.60}$
Mn$_{19.60}$Ti$_{2.00}$
(25~$^\circ$C) &
For the same recorded composition and temperature, UTS ranges from
630 to 2220~MPa and FE from 3\% to 85\%. Processing and testing details are needed to interpret this spread. &
Source/context check required \\

\addlinespace[2pt]

Missing field with property spread &
Al$_{13.33}$Co$_{13.33}$Cr$_{6.67}$
Fe$_{33.33}$Ni$_{33.33}$
(25~$^\circ$C) &
At matched recorded composition and testing temperature, UTS ranges from 766 to 1013~MPa,
while some corresponding FE values are missing, providing directly observable missingness and a property difference requiring additional context. &
Missingness observed; context check required \\

\addlinespace[2pt]

Statistical discrepancy with fitting evidence &
Original literature No.~1
(UTS = 1072~MPa, FE = 55\%) &
The reported FE differs from its neighborhood prediction.
KNN predicts FE $\approx 11.85\%$, while Random Forest also
indicates a clear deviation. &
Model-discrepancy candidate \\

\bottomrule
\end{tabularx}

\tablenote{UTS is ultimate tensile strength (MPa); FE is fracture elongation (\%). Evidence identifies the observed value, missing field or prediction discrepancy; Interpretation states the follow-up review suggested by that evidence.}
}

Engineering elongation has no universal upper bound of 100\%, and its interpretation depends on measurement definition and gauge length. Likewise, processing can alter properties at the same nominal composition and testing temperature. These two candidate types therefore require source and experimental-context checks; fitting residuals help prioritize that review.

For example, in the original literature reference table, sample No.1 reports FE = 55, whereas the KNN-predicted FE is approximately 11.85. Sample No.108 reports UTS = 710 MPa, whereas the KNN-predicted UTS is approximately 1088.69 MPa. These differences direct attention to records whose source values or processing conditions warrant inspection.

\subsection{Historical-reference batch comparison}
The historical experiment tests whether batch scores follow the known contamination order. Table~\ref{tab:memorygat} compares within-batch anchors, historical KNN and trained Memory-GAT over five seeds. All three mean scores decrease with contamination, expressing a consistent average quality trend on their respective score scales.
\inlinetable{tab:memorygat}{Controlled cross-batch historical comparison (five-seed mean $\pm$ SD).}{%
\small
\begin{adjustbox}{max width=\linewidth}
\begin{tabular}{lccc}
\toprule
Contamination & Within-batch anchor & Historical KNN & Memory-GAT \\
\midrule
0\% & 0.8637 $\pm$ 0.0054 & 0.8410 $\pm$ 0.0030 & 0.5912 $\pm$ 0.0663 \\
10\% & 0.8079 $\pm$ 0.0040 & 0.8266 $\pm$ 0.0034 & 0.5872 $\pm$ 0.0635 \\
30\% & 0.7294 $\pm$ 0.0022 & 0.7912 $\pm$ 0.0026 & 0.5776 $\pm$ 0.0583 \\
50\% & 0.6515 $\pm$ 0.0024 & 0.7581 $\pm$ 0.0020 & 0.5657 $\pm$ 0.0520 \\
\bottomrule
\end{tabular}
\end{adjustbox}

\tablenote{Contamination is the injected-error fraction in the current batch. Within-batch anchor is the mean rule/propagation score; Historical KNN averages historical reference scores by inverse-distance weights; Memory-GAT reports $Q_D$. Larger values indicate higher assessed quality within each scoring method. Values are mean $\pm$ sample standard deviation over five seeds.}
}

The within-batch and historical KNN scores preserve the four-batch ordering in all five runs, with mean Spearman correlation 1.00. Memory-GAT preserves the order in four runs and obtains mean correlation 0.96. Historical KNN averages reference scores using descriptor distances, while Memory-GAT learns scores from the reference nodes and their graph connections. The mean scores of both historical methods track the increasing contamination levels.

\subsection{Computational cost}
Dense reliability storage and each bipartite propagation iteration cost $O(nd)$; exact KNN construction has $O(n^2d)$ distance cost. A GAT layer requires $O(nFH+EH)$ operations for input/hidden widths $F,H$ and $E$ edges. Measured with four CPU threads, the 15,999-record workflow takes $7.480\pm0.176$~s and peaks at $699.1\pm0.2$~MiB process RSS; this includes both local GAT comparisons. The 800-anchor Memory-GAT call for 900 current samples takes $0.216\pm0.002$~s with peak process RSS $390.1\pm0.3$~MiB, excluding upstream scoring and anchor selection. Appendix~\ref{app:runtime} reports scaling and environment details. 
\paragraph{Scope of the evaluation.}
The materials benchmark uses controlled input and target errors, with synthetic calibration labels for threshold selection. MACE results compare methods under the same ten-epoch budget, and the historical experiment compares disjoint batches from the same materials source. These settings define the conditions represented by the reported results.

\section{Conclusion}
This work presents a confidence-based framework for materials data quality assessment that integrates domain knowledge, statistical evidence, and graph-structured optimization. 
Rather than treating data quality as a binary anomaly identification problem, the proposed framework formulates quality evaluation as a continuous confidence-modeling problem and establishes an interpretable workflow consisting of nine-dimensional prior construction, knowledge-guided calibration, heterogeneous graph-based confidence propagation, local refinement, and historical-reference batch comparison.

Experiments show that the framework supports both generic anomaly ranking and materials-specific quality assessment. With materials rules disabled, it achieves an average AUC-ROC of 0.866 on 23 ODDS datasets. 
In the materials data cleaning experiment, the identification and removal of low-quality records lead to substantial recovery of downstream model performance, as validated by both conventional machine learning prediction and MACE-based downstream modeling, with rule-based screening increasing mean MACE $R^2$ from $0.590\pm0.053$ to $0.880\pm0.027$ in the repeated subset study. Graph refinement improves mean anomaly-ranking AP from $0.943\pm0.004$ to $0.953\pm0.004$, while rule calibration gives the highest mean screening F1 and test $R^2$ among the evaluated screening configurations. Historical reference scoring tracks batch-quality changes as contamination increases. Together, these results connect continuous confidence, inspectable evidence and task-dependent choice of scoring configuration within one assessment workflow.

Future work will extend graph construction and reference retrieval through approximate search, expand rule coverage using knowledge graphs and literature mining~\citep{yuan2026deep}, and evaluate historical quality anchors across additional materials sources. 

\section*{Code and data availability}

The code and data supporting this study is available at \url{https://github.com/materialsCnicCas/materials-data-quality}.

\section*{Acknowledgements}

This work was supported by the National Key Research and Development Program of China (Grant No. 2025YFE0102600) and the Strategic Priority Research Program of Chinese Academy of Sciences (Grant No.XDA0410404).

\section*{Declarations}

 The authors declare no competing interests.

\bibliographystyle{unsrtnat}

\clearpage

\bibliography{references}

\clearpage
\appendix

\section*{Appendix}
\addcontentsline{toc}{section}{Appendix}

\setcounter{section}{0}
\setcounter{table}{0}
\renewcommand{\theHtable}{appendix.\arabic{table}}
\setcounter{figure}{0}
\setcounter{algorithm}{0}
\renewcommand{\thesection}{A}
\renewcommand{\thetable}{A.\arabic{table}}
\renewcommand{\thefigure}{A.\arabic{figure}}
\renewcommand{\thealgorithm}{A.\arabic{algorithm}}

\section{Supplementary methods and benchmark results}
\label{app:supplement}

\subsection{Complete data quality assessment workflow}
\label{app:workflow}

\begin{revision}
\inlinealgorithm{alg:complete_workflow}
{End-to-end confidence-based materials data quality assessment.}{%
\footnotesize
\setlength{\tabcolsep}{3pt}
\renewcommand{\arraystretch}{0.96}

\begin{tabularx}{\linewidth}{
>{\raggedleft\arraybackslash}p{0.03\linewidth}
>{\raggedright\arraybackslash}X
}
\toprule

\multicolumn{2}{p{0.97\linewidth}}{
\textbf{Input:}
dataset $X$; metadata $M$; feature-quality weights $v$;
rule sets $\mathcal{R}_u$, $\mathcal{R}_m$, and $\mathcal{R}_g$;
coefficients $\alpha$ and $\lambda$;
historical quality memory $\mathcal{H}$.
}\\

\multicolumn{2}{p{0.97\linewidth}}{
\textbf{Output:}
sample confidence $C_{\mathrm{final}}$;
feature confidence $Z_{\mathrm{graph}}$;
rule-trigger evidence $\mathcal{E}$;
memory scores $m$;
dataset-level score $Q_D$.
}\\

\midrule

1 &
Compute feature indicators $q_{j,k}$ and initialize
$Z_{\mathrm{init},j}=\sum_kv_kq_{j,k}$;
set $C_{\mathrm{init},i}=0.2$ for exact duplicate-group members and 1 otherwise.
\\

2 &
Generate univariate boundary, multivariate relationship, and global
distribution evidence:
$\mathcal{E}_u=\mathcal{R}_u(X)$,
$\mathcal{E}_m=\mathcal{R}_m(X)$, and
$\mathcal{E}_g=\mathcal{R}_g(X)$.
\\

3 &
Combine the evidence as
$\mathcal{E}=\mathcal{E}_u\cup\mathcal{E}_m\cup\mathcal{E}_g$,
and derive the penalty coefficients $S_i$, $T_j$, and edge-reliability
matrix $B$.
\\

4 &
Calibrate confidence using
$C_{\mathrm{cal},i}=C_{\mathrm{init},i}S_i$ and
$Z_{\mathrm{cal},j}=Z_{\mathrm{init},j}T_j$.
\\

5 &
Compute $D_s$ and $D_f$ and construct
$P=\left[\begin{smallmatrix}0&D_s^{-1}B\\D_f^{-1}B^\top&0\end{smallmatrix}\right]$;
set $H^{(0)}=H_{\mathrm{cal}}=[C_{\mathrm{cal}};Z_{\mathrm{cal}}]$.
\\

6 &
Iterate
$H^{(t+1)}=\alpha H_{\mathrm{cal}}+
(1-\alpha)PH^{(t)}$
until convergence.
\\

7 &
Split the converged state into
$C_{\mathrm{graph}}$ and $Z_{\mathrm{graph}}$.
\\

8 &
Construct
$U_i=[\,\widetilde{x}_i\odot Z_{\mathrm{graph}}
\parallel C_{\mathrm{graph},i}\,]$,
build the sample KNN graph, and train GAT against $1-C_{\mathrm{cal},i}$ to obtain $r_i$.
\\

9 &
Refine sample confidence as
$C_{\mathrm{final},i}
=\operatorname{clip}
(C_{\mathrm{graph},i}(1-\lambda r_i),0,1)$.
\\

10 &
Select frozen anchors covering the confidence-score range from
$\mathcal{H}$ and combine them with the current samples to construct
the memory graph.
\\

11 &
Apply Memory-GAT to obtain $m_i$ and compute
$Q_D=|D|^{-1}\sum_{i\in D}m_i$.
\\

12 &
\textbf{return}
$C_{\mathrm{final}}$, $Z_{\mathrm{graph}}$,
$\mathcal{E}$, $m$, and $Q_D$.
\\

\bottomrule
\end{tabularx}
}
\end{revision}
\begin{revision}
\subsection{Confidence distributions across stages}\label{app:score_diagnostics}
Table~\ref{tab:gap} reports the mean confidence of clean and injected records, together with their difference. These statistics describe how the score distributions change during calibration and graph processing. A mean difference depends on score scale: multiplying all scores by a positive constant changes the difference without changing their ranking. It is therefore considered alongside AP, F1 and downstream prediction, rather than used alone to compare stages. In a separate 12,000-record diagnostic, a fixed cutoff of 0.4 retains all records after propagation. The main screening experiment instead selects a threshold for each scoring configuration on the calibration partition.
\inlinetable{tab:gap}{Confidence for clean and injected groups under the controlled benchmark (five-seed mean $\pm$ SD).}{%
\small
\begin{adjustbox}{max width=\linewidth}
\begin{tabular}{lccc}
\toprule
Configuration & Clean & Injected & Gap \\
\midrule
Complete rules & 0.8878 $\pm$ 0.0064 & 0.3000 $\pm$ 0.0000 & 0.5878 $\pm$ 0.0064 \\
Rules + propagation & 0.7709 $\pm$ 0.0035 & 0.5686 $\pm$ 0.0013 & 0.2023 $\pm$ 0.0023 \\
Rules + GAT & 0.7477 $\pm$ 0.0101 & 0.2238 $\pm$ 0.0029 & 0.5239 $\pm$ 0.0121 \\
Full graph & 0.6483 $\pm$ 0.0074 & 0.4215 $\pm$ 0.0053 & 0.2268 $\pm$ 0.0116 \\
\bottomrule
\end{tabular}
\end{adjustbox}

\tablenote{Clean and Injected are the mean confidence scores of clean and contaminated records, respectively. Gap is Clean minus Injected on the corresponding score scale. Values are mean $\pm$ sample standard deviation over five seeds. Full graph includes both propagation and GAT.}
}

\subsection{Operational definitions of the feature prior}\label{app:prior_details}
The nine indicators are calculated for each scoring batch using the weights in Section~3.1. In the repeated benchmark, each field name also serves as its metadata description; the declared type is float, and the range is either configured or taken from the batch minimum and maximum. Several indicators are consequently constant: median imputation makes the non-missing fraction one, all retained fields are numerical, and the supplied metadata yield the same documentation scores. The feature uniqueness indicator is set to one because duplicate records are handled separately by the sample prior. The formulas below specify each indicator. For feature $j$, $m_j,M_j,\mu_j,\gamma_j$ and $\mathrm{IQR}_j$ denote the minimum, maximum, mean, pandas sample skewness and interquartile range. The configured range is $[l_j,u_j]$, and $\delta=10^{-9}$ prevents division by zero.
\begingroup\small\setlength{\tabcolsep}{4pt}
\begin{longtable}{p{.20\linewidth}p{.43\linewidth}p{.29\linewidth}}
\caption{Feature indicators: implementation names and operational definitions.}\label{tab:dimensions}\\
\toprule Name / interpretation & Formula in this reference experiment & Operational meaning\\\midrule\endfirsthead
\multicolumn{3}{l}{\small Table \thetable\ continued}\\\toprule Name / interpretation & Formula & Operational meaning\\\midrule\endhead
\bottomrule\endfoot
\texttt{provenance} / documentation availability & $\min(1,0.4I[\mathrm{description}]+0.3I[\mathrm{range}]+0.3I[\mathrm{type}])$. & Documentation completeness from available metadata.\\
\texttt{dataset\_}\allowbreak\texttt{characteristics} / range--center agreement & If an endpoint lies outside the configured range, $R=\max(0,1-(|m-l|+|M-u|)/(u-l+\delta))$; otherwise $R=1$. Set $C=\max(0,1-|\mu-(l+u)/2|/(u-l+\delta))$ and $q=(R+C)/2$. & Measures range agreement and distance from the configured range center.\\
\texttt{uniformity} / numeric compatibility & Fraction of nonmissing observations convertible to a number; zero for an empty field. & Numerical conversion rate; constant for the numerical fields used here.\\
\texttt{metadata\_coupling} / lexical match & \texttt{difflib.SequenceMatcher} ratio between lowercased field name and description. & Lexical agreement proxy for field documentation.\\
\texttt{non\_missing\_cells} / observed-cell fraction & Fraction nonmissing at scoring time. & Adapted completeness; equals one after imputation here.\\
\texttt{non\_duplicate\_rows} / duplicate handling & $q=1$; exact duplicate detection acts through $C_{\mathrm{init}}$ instead. & Fixed at one; sample duplicates are handled by the sample prior.\\
\texttt{un\_skewness} / shape regularity & With $s=|\gamma|$: $1-0.2s$ for $s\leq0.5$; $0.9-0.4(s-0.5)$ for $0.5<s\leq1$; $0.5-0.3(s-1)$ for $1<s\leq2$; $\max(0,0.2-0.1(s-2))$ otherwise. & Piecewise shape score, with a discontinuity at $s=1$ in the implementation.\\
\texttt{accuracy} / numeric regularity & Start at one. Multiply by $\max(0.5,1-1.5(o-0.10))$ only when Tukey-outlier fraction $o>0.10$ and $\mathrm{IQR}>0$. Multiply by $0.8$ if the standard deviation of decimal digit counts exceeds three, or $0.9$ if it exceeds two. & Numerical regularity proxy using precision and outlier frequency.\\
\texttt{target\_correlation} / principal-variation relevance & $\min(1,\max(0.2,1.2|\operatorname{corr}(x_j,\mathrm{PC1})|))$ using unstandardized numerical PCA; $0.5$ for undefined correlation. & Descriptor-only relevance to the principal variation of the batch.\\
\end{longtable}\endgroup
All nine feature scores are saved for each seed using these fixed operational formulas.

\subsection{Convergence and implementation verification}\label{app:convergence}
All entries of the static $B$ are positive, hence the separate degree matrices have nonzero diagonals and the block transition $P$ is row-stochastic. For $F(H)=\alpha H_{\mathrm{cal}}+(1-\alpha)PH$,
\begin{equation}
\|F(H)-F(G)\|_\infty\leq(1-\alpha)\|H-G\|_\infty.
\end{equation}
For $0<\alpha\leq1$, the map has the unique fixed point $H^*=\alpha[I-(1-\alpha)P]^{-1}H_{\mathrm{cal}}$ and converges geometrically. Every update is a convex combination, preserving $[0,1]$. An independent sparse solution of a small system agrees within $2.6\times10^{-9}$; an independently assembled 12,000-row sparse transition agrees within $1.1\times10^{-13}$.

\subsection{Memory-GAT inputs and training}\label{app:memory_details}
Memory-GAT uses 18 descriptors: band gap, energy above hull, space group, three lattice lengths, three angles and nine lattice-matrix entries. Missing values and standardization are handled using medians, means and standard deviations fitted on the historical data. The historical input is $[\operatorname{standardize}_{\mathcal H}(x),2q_{\mathcal H}-1]$, where $q_{\mathcal H}$ is the stored reference confidence. The current input is $[\operatorname{standardize}_{\mathcal H}(x),2a-1]$, where $a$ is the current rule/propagation score. These 19-dimensional vectors are also used to retrieve Euclidean nearest neighbors. The encoder uses hidden width 24, two first-layer attention heads and a sigmoid MLP output. Each current node receives messages from eight historical neighbors, six current neighbors and itself; historical nodes receive only self-loop messages. The loss is
\begin{align}
\mathcal L_{\mathrm{memory}}={}&\frac4{|\mathcal H|}\sum_{u\in\mathcal H}(\widehat q_u-q_u)^2
+\frac1{|\mathcal C|}\sum_{i\in\mathcal C}(\widehat q_i-a_i)^2\nonumber\\
&+\frac{0.1}{|\mathcal E|}\sum_{(u,v)\in\mathcal E}(\widehat q_u-\widehat q_v)^2.
\end{align}
The first two terms fit historical and current reference scores; the third penalizes differences between connected nodes. Adam uses learning rate $10^{-3}$, weight decay $10^{-5}$, dropout 0.05 and gradient clipping at one. Training runs for at most 30 epochs and stops after eight epochs without a lower training loss; the parameters with the lowest loss are restored. Historical KNN uses eight historical neighbors with inverse-distance weights.

\subsection{Controlled corruption and MACE settings}\label{app:corruption}
Within each corrupted batch, 50\% of rows are sampled and divided into four equal groups. Statistical corruption perturbs up to eight generic descriptors by independent Gaussian offsets at $4\sigma$; boundary corruption sets up to five configured fields 35\% of their range below or above the endpoints. Relationship corruption replaces lattice lengths with 2 or 30~\AA\ according to their stored row norm. Mixed corruption perturbs up to 12 descriptors at the declared scale, shifts the first eight descriptors by $1.8\sigma$, may set a negative band gap, and modifies one lattice length. Every selected row receives target noise at $2.8\sigma_y$. The same recipe is applied independently to fitting and calibration batches. Corruption labels are used for metrics and calibration-threshold selection, not scorer inputs or training targets.

\label{app:mace_details}
For MACE, corrupted numerical lattice and coordinate fields are exported as actual ASE structures, and energy per atom is multiplied by atom count. All conditions use the same structural filter, which requires finite coordinates, absolute cell determinant at least $1$~\AA$^3$, condition number at most $10^4$, minimum periodic pair distance at least $1$~\AA, and at most 120 periodic neighbors within 4~\AA. A fitting record is eligible only if both its original and corrupted structures pass these checks. Validation and test records are checked using their clean structures. Random retention is matched to the screened sample count after structural filtering.

The MACE 0.3.15 environment uses Python 3.10.20, PyTorch 2.11.0 with CUDA 12.8 and NVIDIA RTX 5090 GPUs. Additional settings are max\_L=1, atomic-number vocabulary 1--94, validation batch size 64, Adam with AMSGrad, gradient clipping 10 and ReduceLROnPlateau (factor 0.5, patience 10). Energy loss weight is one and force loss weight zero; stage two is disabled. Clean validation selects the exported checkpoint, with no test file passed to training. All 35 runs are evaluated on common clean test records within each seed. Every condition uses the same ten-epoch training budget.

\subsection{Additional sensitivity results}\label{app:sensitivity}
Table~\ref{tab:weights} reports the propagation configuration with thresholds selected on the calibration partition. Each indicator weight is multiplied by 0, 0.8 or 1.2, after which all weights are normalized to sum to one. Thus, \texttt{x0.0} removes an indicator, and \texttt{x0.8}/\texttt{x1.2} reduce/increase its weight by 20\% before normalization. The \texttt{uniform} setting gives all nine indicators equal weight. All other parameters remain fixed. Changing a constant indicator can still change the overall score scale through weight normalization.
\begingroup\small\renewcommand{\arraystretch}{0.9}
\begin{longtable}{lccc}
\caption{Complete feature-weight sensitivity results. Setting identifies the indicator and its weight multiplier (x0.0 removes it); uniform gives equal weights. AP is average precision, F1 measures screening, and $R^2$ evaluates HGBR prediction on clean test data. Values are mean $\pm$ sample standard deviation over five seeds.}\label{tab:weights}\\
\toprule Setting & AP & F1 & $R^2$\\\midrule\endfirsthead
\toprule Setting & AP & F1 & $R^2$\\\midrule\endhead
\bottomrule\endfoot
accuracy\_x0.0 & 0.9032 $\pm$ 0.0056 & 0.7851 $\pm$ 0.2689 & 0.5857 $\pm$ 0.0425 \\
accuracy\_x0.8 & 0.9025 $\pm$ 0.0054 & 0.7726 $\pm$ 0.2766 & 0.5828 $\pm$ 0.0430 \\
accuracy\_x1.2 & 0.9023 $\pm$ 0.0053 & 0.7683 $\pm$ 0.2807 & 0.5858 $\pm$ 0.0371 \\
dataset\_characteristics\_x0.0 & 0.9067 $\pm$ 0.0049 & 0.6881 $\pm$ 0.2692 & 0.5656 $\pm$ 0.0392 \\
dataset\_characteristics\_x0.8 & 0.9040 $\pm$ 0.0052 & 0.7657 $\pm$ 0.2816 & 0.5828 $\pm$ 0.0414 \\
dataset\_characteristics\_x1.2 & 0.9009 $\pm$ 0.0055 & 0.7837 $\pm$ 0.2726 & 0.5866 $\pm$ 0.0427 \\
default & 0.9024 $\pm$ 0.0054 & 0.7715 $\pm$ 0.2776 & 0.5828 $\pm$ 0.0433 \\
metadata\_coupling\_x0.0 & 0.9003 $\pm$ 0.0056 & 0.7827 $\pm$ 0.2714 & 0.5849 $\pm$ 0.0402 \\
metadata\_coupling\_x0.8 & 0.9020 $\pm$ 0.0054 & 0.7720 $\pm$ 0.2776 & 0.5836 $\pm$ 0.0434 \\
metadata\_coupling\_x1.2 & 0.9028 $\pm$ 0.0053 & 0.7688 $\pm$ 0.2798 & 0.5826 $\pm$ 0.0425 \\
non\_duplicate\_rows\_x0.0 & 0.9011 $\pm$ 0.0055 & 0.7813 $\pm$ 0.2715 & 0.5835 $\pm$ 0.0402 \\
non\_duplicate\_rows\_x0.8 & 0.9021 $\pm$ 0.0054 & 0.7716 $\pm$ 0.2777 & 0.5851 $\pm$ 0.0413 \\
non\_duplicate\_rows\_x1.2 & 0.9027 $\pm$ 0.0054 & 0.7713 $\pm$ 0.2777 & 0.5823 $\pm$ 0.0429 \\
non\_missing\_cells\_x0.0 & 0.8980 $\pm$ 0.0058 & 0.8380 $\pm$ 0.2657 & 0.5976 $\pm$ 0.0357 \\
non\_missing\_cells\_x0.8 & 0.9014 $\pm$ 0.0055 & 0.7811 $\pm$ 0.2713 & 0.5839 $\pm$ 0.0376 \\
non\_missing\_cells\_x1.2 & 0.9034 $\pm$ 0.0053 & 0.7675 $\pm$ 0.2809 & 0.5827 $\pm$ 0.0420 \\
provenance\_x0.0 & 0.9011 $\pm$ 0.0055 & 0.7813 $\pm$ 0.2715 & 0.5835 $\pm$ 0.0402 \\
provenance\_x0.8 & 0.9021 $\pm$ 0.0054 & 0.7716 $\pm$ 0.2777 & 0.5851 $\pm$ 0.0413 \\
provenance\_x1.2 & 0.9027 $\pm$ 0.0054 & 0.7713 $\pm$ 0.2777 & 0.5823 $\pm$ 0.0429 \\
target\_correlation\_x0.0 & 0.9038 $\pm$ 0.0052 & 0.8434 $\pm$ 0.2550 & 0.5977 $\pm$ 0.0330 \\
target\_correlation\_x0.8 & 0.9027 $\pm$ 0.0053 & 0.7797 $\pm$ 0.2703 & 0.5848 $\pm$ 0.0414 \\
target\_correlation\_x1.2 & 0.9021 $\pm$ 0.0055 & 0.7686 $\pm$ 0.2825 & 0.5817 $\pm$ 0.0422 \\
un\_skewness\_x0.0 & 0.9052 $\pm$ 0.0051 & 0.7435 $\pm$ 0.3276 & 0.5803 $\pm$ 0.0496 \\
un\_skewness\_x0.8 & 0.9031 $\pm$ 0.0054 & 0.7649 $\pm$ 0.2870 & 0.5821 $\pm$ 0.0414 \\
un\_skewness\_x1.2 & 0.9017 $\pm$ 0.0054 & 0.7817 $\pm$ 0.2685 & 0.5834 $\pm$ 0.0415 \\
uniform & 0.9047 $\pm$ 0.0051 & 0.7615 $\pm$ 0.2786 & 0.5807 $\pm$ 0.0396 \\
uniformity\_x0.0 & 0.8988 $\pm$ 0.0056 & 0.7900 $\pm$ 0.2689 & 0.5863 $\pm$ 0.0443 \\
uniformity\_x0.8 & 0.9016 $\pm$ 0.0055 & 0.7810 $\pm$ 0.2713 & 0.5835 $\pm$ 0.0394 \\
uniformity\_x1.2 & 0.9032 $\pm$ 0.0053 & 0.7679 $\pm$ 0.2805 & 0.5823 $\pm$ 0.0418 \\
\end{longtable}\endgroup

\subsection{Computational scaling}\label{app:runtime}
Local timing uses Python 3.9.7, PyTorch 2.7.1 CPU and scikit-learn 1.6.1 on an AMD Ryzen 5 9500F, with four threads and three fresh-process runs per scale. Total scoring includes both 50-epoch local GAT comparisons and supporting steps, but not reading or corruption generation. Peak resident set size (RSS) measures the memory of the entire process, including the Python runtime and data preparation. The static iteration converges in 38--39 iterations.
\inlinetable{tab:runtime}{Local scoring cost (mean $\pm$ SD over three fresh processes).}{%
\small
\begin{adjustbox}{max width=\linewidth}
\begin{tabular}{lccc}
\toprule
Records & Fields & Scoring (s) & Peak RSS (MiB) \\
\midrule
999 & 43--46 & 0.787 $\pm$ 0.022 & 457.9 $\pm$ 27.7 \\
3,999 & 43 & 1.968 $\pm$ 0.004 & 468.0 $\pm$ 12.0 \\
8,000 & 43 & 3.684 $\pm$ 0.026 & 529.5 $\pm$ 0.7 \\
15,999 & 43 & 7.480 $\pm$ 0.176 & 699.1 $\pm$ 0.2 \\
\bottomrule
\end{tabular}
\end{adjustbox}

\tablenote{Records and Fields give the numbers of samples and numerical descriptors. Scoring is elapsed scoring time in seconds. Peak RSS is the maximum resident set size of the entire process, in mebibytes (MiB). Values are mean $\pm$ sample standard deviation over three independent process runs.}
}

Historical scoring uses 900 current records and 200/400/800 anchors. Mean scorer-call times are 0.205/0.210/0.216~s, with process RSS 388.4/389.0/390.1~MiB. These include descriptor preparation, retrieval, attention training and inference, excluding upstream rule scoring and anchor selection. Dense propagation uses $O(nd)$ storage and per-iteration arithmetic. Exact sample KNN uses $O(n^2d)$ arithmetic; historical/current retrieval has worst-case $O(NM(d+1))$/$O(N^2(d+1))$ distance cost even when KD-tree pruning is used. The timing measurements use exact retrieval on one machine.

\subsection{Alloy audit criteria}\label{app:case_audit}
Missingness counts are computed from the original supplied workbooks before imputation. The 185-record table lacks all 12 processing-history columns listed in Table~\ref{tab:alloy_fields}; the 315-record source table contains them. Candidate groups use the 14 element fractions rounded to four decimals and, in the 185-record table, recorded testing temperature. Missing/nonnumeric fractions are replaced by zero for group-key construction. Each property is examined separately within a group. With at least two observations, the group is flagged for that property when UTS spread is at least 100~MPa and relative spread at least 0.15, or FE spread is at least five percentage points and relative spread at least 0.20. Relative spread is the maximum minus the minimum, divided by the absolute median with a lower bound of $10^{-6}$ on the denominator. These flags prioritize source review using the recorded composition, temperature and property fields.
\end{revision}

\subsection{Field definitions for the alloy case-study datasets}
\label{app:alloy_fields}

The two alloy tables differ primarily in their representation of experimental conditions.
The 185-record case-study table retains a single testing-temperature field, whereas the
315-record source-literature table provides a more detailed processing history.
Both tables contain the same elemental-composition variables, mechanical-property
variables, and 32 composition-derived material descriptors.

\begingroup
\footnotesize
\setlength{\tabcolsep}{3pt}
\renewcommand{\arraystretch}{1.10}
\setlength{\LTpre}{3pt}
\setlength{\LTpost}{3pt}

\begin{longtable}{
>{\raggedright\arraybackslash}p{0.10\linewidth}
>{\raggedright\arraybackslash}p{0.17\linewidth}
>{\raggedright\arraybackslash}p{0.25\linewidth}
>{\raggedright\arraybackslash}p{0.40\linewidth}
}
\caption{Field categories and definitions for the 185-record case-study table and the 315-record source-literature reference table.}
\label{tab:alloy_fields}\\

\toprule
Availability & Category & Fields & Definition \\
\midrule
\endfirsthead

\multicolumn{4}{l}{\footnotesize Table~\thetable\ continued}\\[2pt]
\toprule
Availability & Category & Fields & Definition \\
\midrule
\endhead

\midrule
\multicolumn{4}{r}{\footnotesize Continued on the next page}\\
\endfoot

\bottomrule
\endlastfoot

Both &
Sample identifier &
\texttt{Alloy\_Name} (185-record table);
\texttt{No.} (315-record table) &
Alloy-composition label or numerical record identifier used to distinguish individual observations. \\

Both &
Elemental composition &
\texttt{Co}, \texttt{Cr}, \texttt{Fe}, \texttt{Mn}, \texttt{Ni},
\texttt{Al}, \texttt{Ti}, \texttt{V}, \texttt{Cu}, \texttt{W},
\texttt{Mo}, \texttt{Zr}, \texttt{Nb}, \texttt{Ta} &
Contents of the 14 constituent elements. The values describe the nominal elemental fractions recorded for each alloy, typically in atomic percent. \\

185 only &
Testing condition &
\texttt{temperature} &
Temperature at which the reported UTS and FE values were measured. \\

315 only &
Homogenization &
\texttt{Hom-T}, \texttt{Hom-t} &
Homogenization temperature and holding time, respectively. \\

315 only &
Deformation &
\texttt{Def-T}, \texttt{Def-r} &
Deformation temperature and deformation or reduction ratio, respectively. \\

315 only &
First annealing stage &
\texttt{AnnI-T}, \texttt{AnnI-t} &
Temperature and holding time of the first annealing treatment. \\

315 only &
Rolling &
\texttt{Rol-T}, \texttt{Rol-rate} &
Rolling temperature and rolling-reduction rate, respectively. \\

315 only &
Second annealing stage &
\texttt{AnnII-T}, \texttt{AnnII-t} &
Temperature and holding time of the second annealing treatment. \\

315 only &
Aging &
\texttt{Age-T}, \texttt{Age-t} &
Aging temperature and aging duration, respectively. \\

Both &
Mechanical properties &
\texttt{UTS}, \texttt{FE} &
Ultimate tensile strength and fracture elongation, which characterize alloy strength and ductility, respectively. \\

Both &
Atomic and periodic descriptors &
\texttt{Z}, \texttt{Per}, \texttt{Gro}, \texttt{M}, \texttt{VEC} &
Composition-weighted atomic number, periodic-table period, group number, atomic mass, and valence-electron concentration. \\

Both &
Electronic and chemical descriptors &
\texttt{Aff}, \texttt{Ion}, $\chi$ &
Composition-weighted electron affinity, first ionization energy, and electronegativity. \\

Both &
Basic physical and thermal descriptors &
$\rho$, \texttt{Tm}, \texttt{Tb}, \texttt{k}/$\kappa$, \texttt{Cm} &
Composition-weighted density, melting point, boiling point, thermal conductivity, and molar heat capacity. The thermal-conductivity field is named \texttt{k} in the 185-record table and $\kappa$ in the 315-record table. \\

Both &
Phase-change thermodynamic descriptors &
$\Delta H_{\mathrm{v}}$, $\Delta H_{\mathrm{f}}$, $\Delta S_{\mathrm{f}}$ &
Composition-weighted enthalpy of vaporization, enthalpy of fusion, and entropy of fusion. \\

Both &
Electronic, transport, and structural descriptors &
\texttt{ED}, \texttt{D}, \texttt{a}, \texttt{R} &
Electron-density descriptor, diffusion-related coefficient, lattice parameter, and atomic-radius descriptor supplied by the source dataset. \\

Both &
Elastic descriptors &
\texttt{E}, \texttt{K}, \texttt{G},
\texttt{C11}, \texttt{C12}, \texttt{C44} &
Composition-weighted Young's modulus, bulk modulus, shear modulus, and the three independent elastic stiffness constants commonly used for cubic materials. \\

Both &
Vibrational descriptor &
\texttt{Td} &
Composition-weighted Debye temperature, characterizing the characteristic lattice-vibration scale. \\

Both &
Alloy-level mismatch and thermodynamic descriptors &
$\Delta R$, $\Delta\chi$, $\Delta S_{\mathrm{mix}}$,
$\Delta H_{\mathrm{mix}}$, $\Omega$ &
Atomic-radius mismatch, electronegativity mismatch, configurational entropy of mixing, enthalpy of mixing, and the thermodynamic $\Omega$ parameter combining melting temperature, mixing entropy, and mixing enthalpy. \\

\end{longtable}

\noindent\footnotesize
\textit{Note:} The field names and grouping follow the uploaded source tables.
The units and numerical conventions are retained from the corresponding literature data.
The physicochemical descriptors are primarily composition-weighted elemental properties,
whereas $\Delta R$, $\Delta\chi$, $\Delta S_{\mathrm{mix}}$,
$\Delta H_{\mathrm{mix}}$, and $\Omega$ are alloy-level quantities derived from the overall composition.

\endgroup

\subsection{Detailed ODDS benchmark results}
\label{app:odds}

\inlinetable{tab:odds_full}{Detailed ODDS benchmark comparison in terms of AUC-ROC.}{%
\begin{adjustbox}{width=\linewidth}
\scriptsize
\setlength{\tabcolsep}{2.2pt}
\begin{tabular}{lccccccccccccccc}
\toprule
Dataset & \multicolumn{4}{c}{Classical methods} & \multicolumn{8}{c}{Deep-learning-based methods} & \multicolumn{2}{c}{AnoLLM} & Our method \\
\cmidrule(lr){2-5}\cmidrule(lr){6-13}\cmidrule(lr){14-15}
 & IForest & PCA & KNN & ECOD & DeepSVDD & RCA & SLAD & GOAD & NeuTraL & ICL & DTE & REPEN & 135M & 360M &  \\
\midrule
Annthyroid & 0.922 & 0.839 & 0.811 & 0.790 & 0.742 & 0.718 & 0.761 & 0.572 & 0.813 & 0.842 & 0.977 & 0.736 & 0.927 & 0.931 & 0.924 \\
Arrhythmia & 0.827 & 0.796 & 0.786 & 0.811 & 0.765 & 0.786 & 0.784 & 0.681 & 0.760 & 0.785 & 0.771 & 0.684 & 0.825 & 0.822 & 0.790 \\
BreastW & 0.994 & 0.988 & 0.992 & 0.992 & 0.974 & 0.987 & 0.986 & 0.994 & 0.983 & 0.992 & 0.982 & 0.955 & 0.992 & 0.993 & 0.993 \\
Cardio & 0.948 & 0.966 & 0.921 & 0.935 & 0.842 & 0.948 & 0.840 & 0.524 & 0.859 & 0.894 & 0.920 & 0.829 & 0.940 & 0.873 & 0.926 \\
ForestCover & 0.870 & 0.945 & 0.985 & 0.921 & 0.533 & 0.944 & 0.857 & 0.278 & 0.898 & 0.977 & 0.978 & 0.902 & 0.881 & 0.835 & 0.902 \\
Glass & 0.802 & 0.713 & 0.849 & 0.693 & 0.824 & 0.719 & 0.790 & 0.574 & 0.933 & 0.887 & 0.799 & 0.755 & 0.819 & 0.797 & 0.760 \\
Ionosphere & 0.891 & 0.894 & 0.960 & 0.734 & 0.963 & 0.916 & 0.960 & 0.950 & 0.956 & 0.970 & 0.964 & 0.545 & 0.909 & 0.924 & 0.924 \\
LetterRecognition & 0.631 & 0.529 & 0.865 & 0.567 & 0.776 & 0.716 & 0.908 & 0.811 & 0.929 & 0.959 & 0.872 & 0.597 & 0.967 & 0.867 & 0.766 \\
Lymphography & 0.673 & 0.826 & 0.860 & 0.830 & 0.899 & 0.919 & 0.964 & 0.817 & 0.847 & 0.827 & 0.909 & 0.808 & 0.968 & 0.993 & 0.953 \\
Mammography & 0.881 & 0.900 & 0.872 & 0.906 & 0.857 & 0.873 & 0.740 & 0.756 & 0.690 & 0.782 & 0.864 & 0.863 & 0.915 & 0.876 & 0.884 \\
Musk & 0.971 & 1.000 & 1.000 & 0.956 & 1.000 & 1.000 & 1.000 & 1.000 & 1.000 & 1.000 & 1.000 & 0.722 & 1.000 & 1.000 & 1.000 \\
Optdigits & 0.824 & 0.574 & 0.944 & 0.606 & 0.753 & 0.800 & 0.735 & 0.847 & 0.979 & 0.955 & 0.888 & 0.607 & 0.983 & 0.939 & 0.687 \\
Pendigits & 0.971 & 0.942 & 0.999 & 0.928 & 0.838 & 0.967 & 0.932 & 0.220 & 0.963 & 0.971 & 0.982 & 0.922 & 0.971 & 0.964 & 0.969 \\
Pima & 0.720 & 0.711 & 0.741 & 0.587 & 0.593 & 0.704 & 0.584 & 0.665 & 0.763 & 0.697 & 0.662 & 0.688 & 0.663 & 0.654 & 0.745 \\
Satellite & 0.807 & 0.662 & 0.874 & 0.582 & 0.800 & 0.734 & 0.864 & 0.795 & 0.859 & 0.854 & 0.789 & 0.740 & 0.902 & 0.877 & 0.832 \\
Satimage-2 & 0.993 & 0.979 & 0.999 & 0.965 & 0.961 & 0.998 & 0.997 & 0.994 & 0.861 & 0.997 & 0.988 & 0.998 & 1.000 & 0.999 & 0.995 \\
Shuttle & 0.996 & 0.994 & 0.999 & 0.993 & 0.997 & 0.996 & 0.998 & 0.990 & 0.997 & 0.999 & 0.998 & 0.992 & 1.000 & 1.000 & 0.993 \\
Speech & 0.478 & 0.471 & 0.486 & 0.471 & 0.563 & 0.472 & 0.511 & 0.550 & 0.609 & 0.582 & 0.511 & 0.531 & 0.470 & 0.470 & 0.615 \\
Thyroid & 0.989 & 0.984 & 0.976 & 0.978 & 0.869 & 0.969 & 0.948 & 0.689 & 0.886 & 0.987 & 0.990 & 0.904 & 0.975 & 0.983 & 0.985 \\
Vertebral & 0.446 & 0.494 & 0.406 & 0.474 & 0.478 & 0.478 & 0.483 & 0.516 & 0.545 & 0.543 & 0.570 & 0.247 & 0.565 & 0.408 & 0.403 \\
Vowels & 0.779 & 0.644 & 0.975 & 0.597 & 0.895 & 0.891 & 0.969 & 0.925 & 0.988 & 0.979 & 0.983 & 0.753 & 0.982 & 0.938 & 0.946 \\
WBC & 0.947 & 0.949 & 0.947 & 0.907 & 0.904 & 0.946 & 0.928 & 0.663 & 0.765 & 0.908 & 0.913 & 0.770 & 0.964 & 0.952 & 0.944 \\
Wine & 0.930 & 0.927 & 0.952 & 0.729 & 0.854 & 0.899 & 0.964 & 0.961 & 0.960 & 0.944 & 0.974 & 0.835 & 0.909 & 0.851 & 0.973 \\
\midrule
Average & 0.839 & 0.814 & 0.878 & 0.781 & 0.812 & 0.843 & 0.848 & 0.729 & 0.863 & 0.884 & 0.882 & 0.756 & 0.892 & 0.867 & 0.866 \\
Rank & 10 & 11 & 4 & 13 & 12 & 9 & 8 & 15 & 7 & 2 & 3 & 14 & 1 & 5 & 6 \\
\bottomrule
\end{tabular}
\end{adjustbox}

\tablenote{All entries are AUC-ROC (area under the receiver operating characteristic curve); higher values indicate better anomaly ranking. IForest, PCA and KNN denote Isolation Forest, principal component analysis and nearest-neighbor detection. AnoLLM headings 135M and 360M identify model sizes in millions of parameters. Other column names identify the baseline models described in the benchmark comparison.}
}

\end{document}